\documentclass[11pt]{article}

\usepackage[utf8]{inputenc}
\usepackage[T1]{fontenc}
\usepackage{lmodern}
\usepackage{amsmath, amssymb}
\usepackage{graphicx}
\usepackage{hyperref}
\usepackage{booktabs}
\usepackage{multirow}
\usepackage{subcaption}
\usepackage{siunitx}\usepackage{authblk}

\usepackage{geometry}
\newcommand{\PSABench}{PSA-Bench2019}
\newcommand{\SVATest}{SVA-Test}

\title{FarSky: Task-Aware Latent-Space Coupling for Generative Intra-Hour Solar Forecasting}

\author[1,2]{Yann Fabel\thanks{Corresponding author.}}
\author[1]{Bijan Nouri}
\author[1,2]{Milon Miah}
\author[1]{Niklas Blum}
\author[3]{Luis F. Zarzalejo}
\author[2]{Julia Kowalski}
\author[1,2]{Robert Pitz-Paal}

\affil[1]{DLR, Institute of Solar Research}
\affil[2]{RWTH Aachen University}
\affil[3]{CIEMAT, Energy Department, Renewable Energy Division}

\date{}

\begin{document}

\maketitle

\begin{abstract}
Accurate solar irradiance forecasting is essential for the reliable integration of photovoltaic power into modern electricity grids. 
All-sky imagers (ASI) provide high-resolution observations of clouds, making them well suited for intra-hour forecasting.
Recent deep learning approaches have substantially improved forecast accuracy but are often limited by deterministic predictions and a reduced capability to anticipate ramp events. 
This work proposes FarSky, a generative forecasting framework that leverages latent-space coupling to learn task-aware representations of sky images.
A multi-task autoencoder first learns a shared latent representation for image reconstruction and irradiance estimation.
A latent diffusion model then generates future latent states conditioned on recent observations, from which irradiance forecasts are directly decoded. 
Probabilistic forecasts are inherently obtained through stochastic sampling.
The framework is developed using a multi-year ASI dataset acquired at the Plataforma Solar de Almería, Spain, and evaluated on two independent test datasets against persistence, state-of-the-art end-to-end, and generative forecasting approaches. 
FarSky achieves the best overall deterministic and probabilistic forecasting performance, improving forecast skill by up to 11 percentage points. 
Furthermore, it substantially improves ramp event detection over existing methods, achieving F1-scores above 60\%.
These results demonstrate the potential of combining generative models with task-aware latent-space coupling for solar forecasting.

\end{abstract}

\section{Introduction}

Achieving climate neutrality by mid-century, as targeted by the Paris Agreement and the European Green Deal, requires a rapid expansion of renewable energy sources to decarbonize the electricity sector.
Among these, solar photovoltaics (PV) has emerged as one of the fastest-growing technologies worldwide, driven by substantial cost reductions and large-scale deployment.
Global installed PV capacity increased from about 41\,GW in 2010 to about 3\,TW by the end of 2025~\cite{IEAPVPS2026_Fact}, with projections exceeding 7\,TW by 2030~\cite{Masson2025_Snapshot,SolarPowerEurope2025_Global}.
As PV penetration continues to increase, maintaining the reliable operation of electricity grids becomes increasingly challenging due to the inherent variability of solar power generation.

While deterministic components of the solar resource, such as the diurnal and seasonal cycles, can be predicted accurately, cloud-induced irradiance fluctuations remain difficult to forecast.
Rapid cloud movements can cause substantial changes in solar irradiance within seconds to minutes, resulting in sudden variations in PV power output known as ramp events.
These fluctuations may lead to voltage and frequency instabilities, increase the need for fast-reacting reserve capacity or energy storage, and complicate participation in electricity markets~\cite{Perez2016_Spatial,Chu2021_Intra,Nouri2023_Probabilistic,Paletta2023_Advances}.
At the same time, the highly nonlinear and chaotic evolution of cloud fields on these spatial and temporal scales makes the occurrence, timing, and magnitude of such fluctuations inherently difficult to predict, resulting in substantial forecast uncertainty.
Consequently, accurate intra-hour solar forecasting is increasingly recognized as an important enabling technology for power system operation, predictive control of storage systems, and the integration of large shares of solar energy into future electricity grids~\cite{West2014_Short,Yang2022_Concise}.

Forecasts on intra-hour time scales require observation systems capable of resolving cloud dynamics at high spatial and temporal resolution.
Ground-based All-Sky Imagers (ASIs) have therefore become a central sensing technology for solar nowcasting.
Unlike satellite observations and numerical weather prediction models, which are limited by comparatively coarse temporal and spatial resolutions, ASIs continuously observe the complete sky dome at sub-minute cadence and provide detailed information about local cloud evolution.
This makes them particularly suitable for predicting cloud motion and the resulting irradiance variability over forecasting horizons within the intra-hour range.

Driven by recent advances in computer vision, deep learning has become the dominant paradigm for ASI-based solar forecasting.
Compared to traditional cloud segmentation and motion estimation techniques, data-driven models can directly learn complex, nonlinear cloud dynamics from image sequences without relying on handcrafted features or simplified physical assumptions.
Numerous end-to-end forecasting models have demonstrated excellent performance in terms of aggregated error metrics by directly predicting future irradiance or PV power from recent observations~\cite{Sun2018_Solar,Fabel2024_Combining}.
However, because these models are typically optimized using regression losses such as the mean squared error, they tend to produce overly smooth forecasts that underestimate rapid irradiance fluctuations and fail to accurately represent steep ramp events~\cite{Paletta2021_Benchmarking}.

To better preserve cloud dynamics during forecasting, recent research has shifted towards approaches that explicitly predict future sky images before deriving irradiance. 
Since local irradiance variations are primarily caused by clouds passing in front of the sun, predicting future cloud evolution provides a physically meaningful intermediate representation for solar forecasting. 
Early work by Le Guen et al.~\cite{LeGuen2020_Deep} adapted the PhyDNet video prediction architecture to forecast future sky images while explicitly modeling cloud motion through learned physical dynamics.
Subsequently, Paletta et al.~\cite{Paletta2022_ECLIPSE} proposed the ECLIPSE framework, which jointly learns spatio-temporal representations for cloud segmentation and irradiance prediction, demonstrating the benefits of task-aware latent representations for downstream forecasting.
More recently, Nie et al.~\cite{Nie2024_SkyGPT} introduced SkyGPT, a generative transformer-based model that predicts future sky images and derives probabilistic irradiance forecasts from the generated image sequences.
Although these approaches demonstrate the advantages of predicting future cloud evolution, they also highlight complementary limitations.
End-to-end frameworks such as PhyDNet and ECLIPSE rely on deterministic future-state prediction, whereas current generative models like SkyGPT primarily learn latent representations optimized for image synthesis rather than downstream irradiance estimation.
Moreover, existing video prediction models continue to exhibit reduced image fidelity at longer lead times, including blurred cloud structures, loss of fine-scale details, and decreased physical consistency.

Recent advances in computer vision suggest diffusion models as a promising foundation for generative solar forecasting.
Diffusion-based models have established the state-of-the-art in image and video synthesis by producing sharp, photorealistic, and temporally coherent predictions~\cite{Rombach2022_High, Blattmann2023_Stable}.
Unlike previous generative architectures, they iteratively refine samples through a denoising process~\cite{Ho2020_Denoising}, enabling the generation of detailed cloud structures while naturally representing uncertainty through stochastic sampling.
Combining these generative capabilities with task-aware latent representations offers the potential to improve both the visual quality of predicted sky images and the accuracy of the derived irradiance forecasts.

In this work, we propose \textit{FarSky}, a diffusion-based framework for generative intra-hour solar forecasting from all-sky imagery.
The proposed approach combines a task-aware latent representation, jointly optimized for image reconstruction and irradiance estimation, with a latent diffusion model for future sky image prediction.
Future irradiance trajectories are obtained directly from the generated sky images, while repeated sampling enables probabilistic forecasting without requiring separate uncertainty models.
The proposed framework is evaluated on multiple years of ASI observations and compared with state-of-the-art deterministic and generative forecasting approaches.
Our experiments demonstrate that task-aware latent-space learning substantially improves irradiance prediction compared to decoupled latent video prediction while simultaneously providing realistic future sky images and competitive probabilistic forecasts.

The remainder of this paper is organized as follows.
Section~\ref{sec:data} describes the datasets and preprocessing procedures.
Section~\ref{sec:methodology} presents the proposed forecasting framework.
Section~\ref{sec:exp_setup} details the experimental setup and the comparison methods.
Section~\ref{sec:results} presents the experimental results.
Finally, Section~\ref{sec:conclusion} summarizes the main findings and discusses directions for future research.

\section{Data}
\label{sec:data}
This section describes the data used throughout this study. 
It first introduces the measurement site and instrumentation, followed by the preprocessing procedures applied to construct the datasets for model development and evaluation.

\subsection{Measurement Site and Instrumentation}
All image and irradiance data used in this study were acquired at the Plataforma Solar de Almería (PSA), a solar research facility owned and operated by CIEMAT in Tabernas, southern Spain. 
The measurement site is located at 37.09$^\circ$N, 2.35$^\circ$W and an altitude of approximately 500\,m above sea level. 
According to the Köppen climate classification, the region is characterized as a cold desert climate (BWk), resulting in predominantly clear-sky conditions throughout the year. 
Nevertheless, the surrounding mountain ranges, including the Sierra Nevada, Sierra Filabres, and Sierra Alhamilla, frequently induce complex cloud formations, rapid cloud evolution, and multilayer cloud structures. 
These characteristics make the site well suited for developing and evaluating intra-hour solar forecasting models.

\begin{figure}
    \centering
    \includegraphics[width=\linewidth]{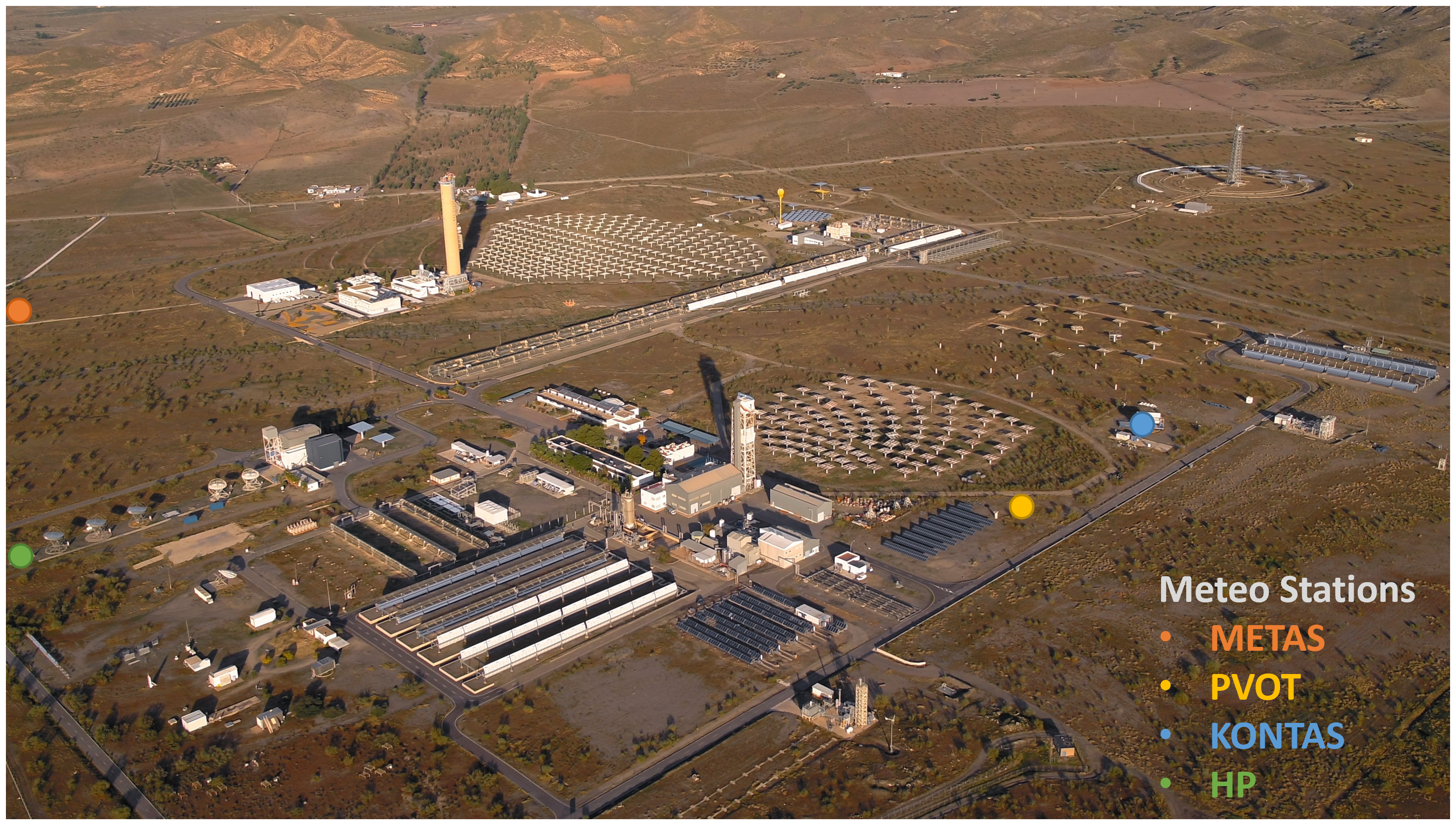}
    \caption{Aerial view of the Plataforma Solar de Almería (PSA) indicating the locations of the utilized measurement stations for model development and evaluation.}
    \label{fig:overview_psa}
\end{figure}

The datasets used in this study comprise image and irradiance measurements acquired at four meteorological stations within the PSA (HP, KONTAS, METAS, and PVOT) between 2019 and 2024. 
Each station is equipped with an all-sky imager and collocated radiometers. 
An overview of the research facility and the location of the respective measurement stations is given in Figure~\ref{fig:overview_psa}.
Data from the PVOT station corresponds to the publicly available SolarVision Almería (SVA) dataset~\cite{Fabel2026_comprehensive}, whereas the remaining measurements originate from earlier and complementary data acquisition at PSA.
Combining observations from multiple stations and acquisition periods substantially increases the diversity of atmospheric conditions, camera viewpoints, and seasonal variability available for model development.

Sky images were acquired by commercial off-the-shelf Mobotix Q25 and Q26 RBG camera systems equipped with fisheye lenses, capturing images with a resolution of $2048\times2112$ pixels. 
Both camera models employ the same CMOS sensor and were operated with a fixed exposure time of 160\,\si{\micro\second}, resulting in highly consistent image characteristics across the datasets produced. 
Images were recorded every 30\,s during daytime. 

At each measurement station, Global Horizontal Irradiance (GHI) was measured by a collocated ISO 9060:2018 Class~A Kipp \& Zonen pyranometer. 
Irradiance was sampled at 1\,Hz, from which 1-minute averages were computed, with each average timestamp corresponding to the end of the respective averaging interval.

\subsection{Data Preprocessing}

Prior to dataset construction, all image and irradiance measurements underwent a common preprocessing procedure to ensure consistency across the different measurement stations and camera systems.

Quality control of the irradiance measurements followed the screening and flagging procedure proposed by Geuder et al.~\cite{Geuder2015_Screening}, combining automated consistency checks with manual inspection to identify physically implausible measurements and periods affected by maintenance or sensor failures.

To synchronize image and irradiance measurements, the 1-minute irradiance averages were linearly interpolated to the corresponding image timestamps.
Since image acquisition was not always perfectly aligned with the full- or half-minute irradiance measurements, a temporal tolerance of $\pm10$\,s relative to the nominal sampling times was allowed to ensure a regular temporal sampling grid for the interpolated irradiance values.

Prior to model training, all sky images were processed using the same preprocessing pipeline. 
First, a geometric camera calibration~\cite{Blum2025_Geometric} was used to determine the zenith position, allowing each fisheye image to be cropped to a square region with the zenith located at the image center and the horizon approximately aligned with the image borders. 
The cropped images were subsequently resized to a spatial resolution of $128\times128$ pixels. 
No camera-specific masking, geometric undistortion, or additional radiometric normalization was applied, allowing the models to directly learn from the original fisheye image geometry.

\subsection{Experimental Datasets}
\label{sec:dataset_construction}

The experimental datasets were constructed from the acquired data described in the previous sections. 
To ensure high-quality training and evaluation data, several filtering steps were applied prior to splitting the data into development and evaluation datasets.

First, only samples with solar elevation angles exceeding 10$^\circ$ were retained. 
At lower solar elevations, the quality of the all-sky images deteriorates due to reduced illumination, pronounced fisheye distortion near the horizon, and the limited spatial resolution of distant cloud structures. 
These factors substantially reduce the visual information available for forecasting, while irradiance measurements at low solar elevations are also subject to increased uncertainty.
Furthermore, days dominated almost entirely by clear-sky conditions were excluded through manual inspection of the daily GHI, DNI, and DHI time series.
Since clear-sky conditions are highly prevalent at the measurement site, this filtering step increases the relative proportion of cloud-induced irradiance variability and therefore the information content of the training data.

The remaining data were subsequently partitioned on a per-day basis to avoid temporal leakage between the individual datasets.
Since all stations are located within the Plataforma Solar de Almería and therefore observe the same prevailing atmospheric conditions, the train, validation, and test splits were defined consistently at the site level.
Consequently, whenever observations from multiple stations were available on a given day, data from all corresponding camera--pyranometer pairs were assigned to the same subset.

To obtain an independent test set covering a broad range of cloud conditions, 60 days between August~2022 and July~2024 from the PVOT station of the SVA dataset were selected manually and withheld from the development dataset beforehand.
This dataset will be referred to as \SVATest{} in the remainder of this work.
The selection was based on a stratification according to the DNI variability classes proposed by~\cite{SchroedterHomscheidt2018_Classifying} and visual inspection of the daily irradiance measurements. 
From the remaining PVOT data, 12 additional days distributed throughout 2023 were assigned to the validation set, while all remaining selected data from PVOT, together with the corresponding observations from the HP, KONTAS, and METAS stations were used for model training.

In addition to the \SVATest{} set, model performance is evaluated on the benchmark dataset introduced by ~\cite{Logothetis2022_Benchmarking}, publicly available on Zenodo~\cite{Fabel2026_PSABenchmark2019}.
This benchmark dataset consists of 28 manually selected days acquired at the METAS station that has been used in various previous works for evaluating ASI-based solar nowcasting systems~\cite{Logothetis2022_Benchmarking, Nouri2022_Hybrid, Fabel2024_Combining}. 
Using this benchmark data enables direct comparison with previously published methods under identical evaluation conditions and therefore complements the assessment on the newly presented test set from the public SVA dataset.

An overview of the resulting datasets from the different stations and acquisition periods is provided in Table~\ref{tab:data_overview}.

\begin{table}[t]
\centering
\caption{Overview of the image and irradiance datasets used in this work. All stations provide all-sky images from Mobotix cameras and collocated irradiance measurements. The numbers indicate the number of days assigned to the training, validation, and test sets.}
\label{tab:data_overview}
\begin{tabular}{llcccc}
\toprule
Station & Camera & Acquisition Period & Train & Val. & Test \\
\midrule
HP      & Q26 & Jul.~2019 -- Dec.~2021 & 679  & -- & -- \\
KONTAS  & Q25 & Jul.~2019 -- Mar.~2022 & 746  & -- & -- \\
METAS   & Q25 & Jul.~2019 -- Jul.~2024 & 1219 & -- & 28 \\
PVOT    & Q26 & Aug.~2022 -- Jul.~2024 & 379  & 12 & 60 \\
\bottomrule
\end{tabular}
\end{table}

\section{Methodology}
\label{sec:methodology}

We propose FarSky, a generative forecasting framework for intra-hour solar forecasting based on all-sky imagery. 
The framework consists of a two-stage pipeline combining latent-space video prediction with irradiance estimation.
First, input sky images are encoded into compact latent representations using an autoencoder. 
A diffusion-based video prediction model then generates multi-step forecasts of future latent states, from which future sky images and irradiance estimates  can be derived.

In contrast to existing approaches such as SkyGPT~\cite{Nie2024_Sky}, which employ a separately trained downstream model on reconstructed future images for irradiance estimation, FarSky derives irradiance forecasts directly from the predicted latent representations.
To this end, the proposed framework extends the autoencoder with an additional irradiance decoder head and jointly learns image reconstruction and irradiance prediction in a shared latent space.
This task-aware latent-space coupling encourages the latent representations to preserve not only visual reconstruction features but also irradiance-relevant information.

\begin{figure}
    \centering
    \includegraphics[width=\linewidth]{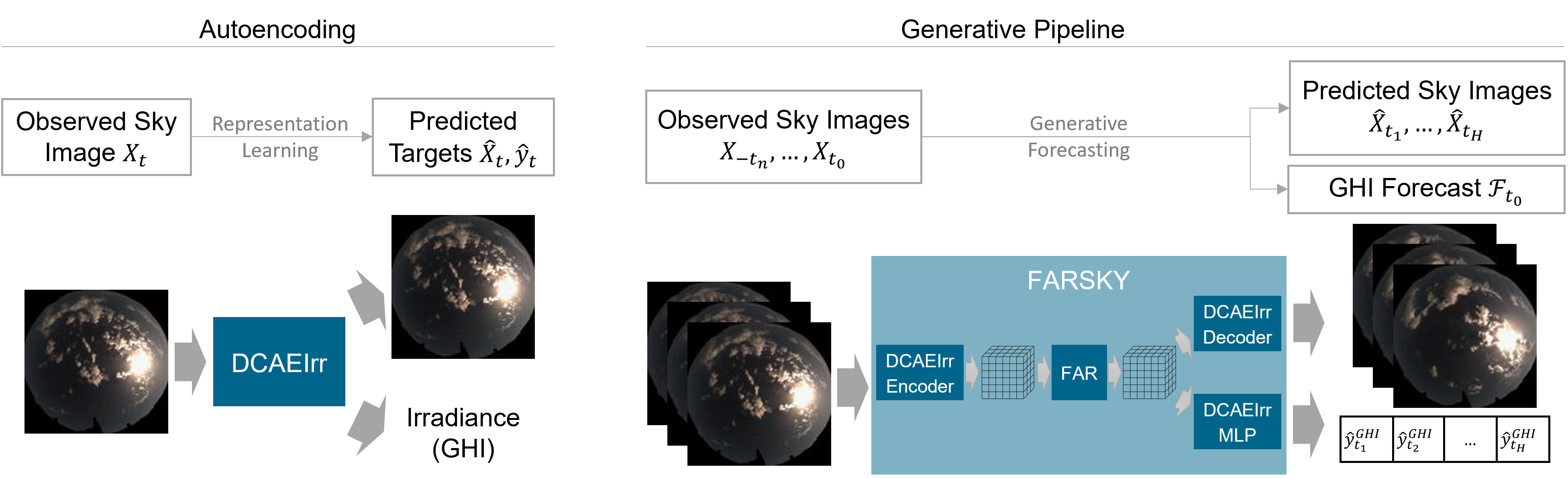}
    \caption{Overview of the proposed FarSky framework. During representation learning (left), the task-aware DCAEIrr autoencoder is trained to jointly reconstruct sky images and estimate GHI. During generative forecasting (right), the image encoder and decoder of the trained DCAEIrr model are reused to map between image and latent space, while the FAR model predicts the future latent representations. Predicted sky images are decoded from the forecast latent states, and the shared irradiance prediction head (MLP) produces the corresponding GHI forecasts.}
    \label{fig:farsky_arch}
\end{figure}

For comparison, a conventional decoupled formulation is additionally evaluated using the identical video prediction backbone together with a separately trained irradiance regression model operating on reconstructed future sky images.
An overview of the proposed framework is shown in Figure~\ref{fig:farsky_arch}.

\subsection{Task-Aware Representation Learning}

The proposed task-aware latent representations are learned by extending the autoencoder used in the original diffusion-based video prediction architecture FAR~\cite{Gu2025_Long}. 
FAR employs a Deep Compression Autoencoder (DCAE), which was specifically designed for efficient latent representations in high-resolution diffusion models~\cite{Chen2024_Deep}. 
In the proposed variant, DCAEIrr, the original encoder-decoder architecture for image compression and reconstruction is retained, while an additional multi-layer perceptron (MLP) head is introduced to predict irradiance directly from the encoded latent representation. 
Given an input sky image $x$, the encoder $E$ produces a latent representation $z=E(x)$, from which both the reconstructed image $\hat{x}$ and the corresponding irradiance estimate $\hat{y}$ are obtained using dedicated decoder heads $D_x, D_y$:
\begin{align*}
\hat{x} &= D_x(z), \\
\hat{y} &= D_y(z).
\end{align*}

To support this multi-task learning objective, the original DCAE loss is augmented with an irradiance regression term. 
The irradiance prediction head is optimized using an L1 loss between predicted and measured irradiance values,

\begin{align*}
\mathcal{L}_{\mathrm{irr}}
&=
\left\lVert
\hat{y}-y
\right\rVert_1, \\
\mathcal{L}
&=
\mathcal{L}_{\mathrm{DCAE}}
+
\lambda_{\mathrm{irr}}
\mathcal{L}_{\mathrm{irr}},
\end{align*}

where $y$ denotes the measured irradiance, $\mathcal{L}_{\mathrm{DCAE}}$ is the original DCAE training objective consisting of pixel-wise reconstruction, perceptual, and adversarial loss components~\cite{Chen2024_Deep}, and $\lambda_{\mathrm{irr}}$ controls the relative contribution of the irradiance regression loss. 
By jointly optimizing image reconstruction and irradiance prediction, the latent representation is encouraged to preserve not only information required for faithful image reconstruction but also features relevant for irradiance estimation.

\subsection{Generative Forecasting in Latent Space}

The learned task-aware latent representations serve as the state space for the subsequent forecasting model based on the Frame AutoRegressive (FAR) diffusion architecture proposed by Gu et al.~\cite{Gu2025_Long}. 
Given a sequence of observed sky images, the encoder of DCAEIrr first maps each frame to its corresponding latent representation. 
These latent representations are subsequently used to train the FAR model, which learns to generate future latent states conditioned on an observed context sequence.

Importantly, no modifications are made to the original FAR architecture or training procedure. 
The proposed approach differs from the original formulation through the use of the task-aware latent representation learned by DCAEIrr. 
This design enables the effect of latent-space coupling to be isolated from changes in the forecasting model itself.

During training, the original FAR training procedure is adopted without modification. Future latent representations are progressively corrupted with Gaussian noise, and the model is optimized using the original flow-matching objective. The loss is computed only over the predicted future latent representations, while the clean latent context frames are excluded.

At inference time, future latent representations are generated autoregressively and can subsequently be decoded either into future sky images using the image decoder or directly into irradiance forecasts using the irradiance MLP head introduced in the previous subsection.
By repeatedly conditioning on previously generated latent states, the model can produce forecasts beyond the forecast horizon used during training, although forecast uncertainty and error generally accumulate with increasing lead time. 
Owing to the stochastic diffusion process, multiple plausible future trajectories can be sampled from the same conditioning sequence.

\subsection{Decoupled Baseline Formulation}

To assess the impact of task-aware latent-space coupling, a decoupled forecasting formulation is considered as a reference. 
In contrast to FarSky, which derives irradiance forecasts directly from the predicted latent representations, the decoupled approach estimates irradiance from reconstructed future sky images using a separately trained downstream model.

The forecasting pipeline is based on the original FAR architecture and employs the unmodified DCAE for latent representation learning. 
Given a sequence of observed sky images, future latent states are generated using the same video prediction model described in the previous subsection and subsequently decoded with the decoder part of the DCAE into synthetic future sky images.

To derive irradiance forecasts, a convolutional neural network (CNN) is trained independently of the generative forecasting model. 
The network takes a single sky image as input and predicts the corresponding clear-sky index of GHI.
Compared to directly regressing GHI, the physically normalized CSI target resulted in more stable optimization for this standalone CNN.
During inference, the trained CNN regressor is applied to the synthetic sky images generated by the FAR video prediction model.
Subsequently, the predicted CSI values are converted back to GHI using clear-sky irradiance obtained from the Ineichen clear-sky model~\cite{Ineichen2002_new} with daily minimum Linke turbidity values, following~\cite{Fabel2026_comprehensive}.
Training is performed exclusively on real sky images and measured irradiance observations, while no temporal context is provided to the model. 

\subsection{Uncertainty Quantification from Ensemble Forecasts}
\label{sec:uncertainty_quantification}

Probabilistic irradiance forecasts are obtained from ensembles of stochastic forecast trajectories. 
For a given input sequence, the diffusion model is sampled multiple times using different noise realizations. 
Each sampled latent trajectory is subsequently converted into an irradiance forecast, yielding an ensemble
\[
\left\{\hat{y}_{h}^{(s)}\right\}_{s=1}^{S}
\]
for each lead time $h$, where $S$ denotes the number of generated trajectories.

Due to the computational cost of diffusion-based sampling, only a limited number of ensemble members can be generated in practice. 
To obtain robust predictive quantiles despite the small ensemble size, a Student-$t$ distribution is estimated independently for each lead time based on the ensemble forecasts. 
The predictive distribution is characterized by the ensemble mean $\mu_h$, standard deviation $\sigma_h$, and degrees of freedom $\nu=S-1$. 
Quantiles are then obtained as
\[
q_h(\tau)
=
\mu_h
+
t_{\tau,\nu}\sigma_h,
\]
where $t_{\tau,\nu}$ denotes the $\tau$-quantile of the Student-$t$ distribution with $\nu$ degrees of freedom. 
Prediction intervals and arbitrary quantiles can subsequently be derived analytically from the fitted distributions.

\section{Experimental Setup}
\label{sec:exp_setup}
This section describes the experimental setup used throughout this work, including model training, comparison methods, and the evaluation protocol. All models are trained and evaluated using the datasets and data splits introduced in Section~\ref{sec:data}.

\subsection{Training Configuration}

All experiments were conducted using the development dataset described in Section~\ref{sec:dataset_construction}. 
The training set was used for parameter optimization, while the validation set served for monitoring convergence and selecting the final model checkpoint.

Following the two-stage training procedure described in Section~\ref{sec:methodology}, the DCAEIrr and FAR models were optimized sequentially. 
For the DCAEIrr, relative weighting factors of the perceptual, discriminator, and irradiance loss weights were set to 1.0, 0.5, and 0.1, respectively.
To stabilize adversarial training, the discriminator was activated only after an initial warm-up period of 100,000 iterations.
To stabilize training, adversarial optimization was enabled only after an initial warm-up period of 100,000 iterations.
Although the DCAEIrr operates on individual images and therefore does not require temporal context during optimization, image sequences were used during training and validation. 
This design choice enables the direct computation of sequence-based image quality metrics and facilitates a consistent comparison with the diffusion-based forecasting model. 
Detailed training settings are shown in in Appendix~\ref{sec:farsky_hyperparameters}.

After training the DCAEIrr, the encoder weights were frozen and used to obtain latent representations for the second optimization stage. 
The diffusion-based forecasting model was trained using the original FAR training configuration and the corresponding hyperparameter settings.
Training was performed using sequences of 16 context frames corresponding to $t_{-15}, \ldots, t_0$ and 16 target frames corresponding to $t_{+1}, \ldots, t_{+16}$, resulting in a temporal resolution of one minute.
A detailed overview of hyperparameters and optimization settings for the video prediction training is provided in Appendix~\ref{sec:farsky_hyperparameters}.

Probabilistic forecasts were generated throughout this work, using an ensemble of $S=4$ stochastic trajectories for each forecast initialization.
This number represents a compromise between computational cost and the ability to estimate predictive uncertainty.
Each trajectory was obtained by independently sampling from Gaussian noise while conditioning on the same sequence of observed sky images.

The forecast horizon was set to 30\,min with a one minute resolution, covering longer time spans compared to the training setup. 
This value was chosen to analyze whether the model still generates reasonable forecasts beyond the training forecast horizon and to provide a higher value for final end users of the forecasts.

To obtain probabilistic irradiance forecasts, each generated sequence of latent representations was processed independently by the irradiance prediction head of the DCAEIrr. 
Forecast quantiles and prediction intervals were subsequently estimated from the resulting ensemble distribution.

\subsection{Comparison Methods}
\label{sec:comparison_methods}

To assess the effectiveness of the proposed latent-coupled forecasting approach, FarSky is compared against four reference methods. 
First, the decoupled generative forecasting formulation (Section~\ref{sec:decoupled_baseline}) is considered to isolate the impact of the proposed latent-space coupling strategy.
Second, Scaled Persistence serves as a standard operational benchmark. 
Third, a transformer-based end-to-end forecasting model is included to compare generative and direct forecasting paradigms. 
Finally, the recently proposed SkyGPT~\cite{Nie2024_SkyGPT} model is used as a state-of-the-art literature baseline.

\subsubsection{Decoupled Baseline}
\label{sec:decoupled_baseline}
The decoupled baseline follows the original video prediction task, in which the DCAE and FAR models are trained exclusively on sky image sequences. 
Irradiance prediction is subsequently decoupled from the video prediction task by estimating irradiance from the synthetic images using a separately trained CNN regressor. 
To ensure a fair comparison, the video prediction models were trained using the same development dataset and optimization settings as employed for FarSky.
A ResNet34-based architecture was employed as CNN regressor, and the complete training configuration is provided in Appendix~\ref{sec:decoupled_hyperparameters}.

\subsubsection{Scaled Persistence}
Scaled Persistence is widely used as a reference method in solar forecasting and serves as a standard baseline for intra-hour forecast horizons\cite{Sengupta2024_Best}. 
The method assumes persistence of the current clear-sky index and computes future irradiance values according to

\begin{equation}
\hat{y}(t+\Delta t) = k_c(t)\,y_{\mathrm{cs}}(t+\Delta t),
\end{equation}

where $k_c(t)$ denotes the clear-sky index at the current time and $y_{cs}$ is the clear-sky irradiance obtained from a clear-sky model.

\subsubsection{End-to-End Deep Learning Baseline} 
As an additional reference, we evaluate a deterministic end-to-end deep learning model (E2E-TF) presented in~\cite{Fabel2024_Combining} that directly predicts multi-step irradiance forecasts without an intermediate generative forecasting stage.
In contrast to the generative approaches, which first generate future sky states before deriving irradiance estimates, the end-to-end model maps historical observations directly to future irradiance values. 
The model combines two transformer-based branches to process complementary information sources. 
A time-series encoder receives recent measurements of GHI, DNI, DHI, and solar position, while a video transformer extracts spatio-temporal features from sequences of all-sky images. 
The resulting feature representations are fused and processed by a multilayer perceptron to predict irradiance values for all lead times within the forecasting horizon. 

To enable comparison with the probabilistic generative approaches, deterministic forecasts are converted into probabilistic forecasts using the quantile-based post-processing method proposed in~\cite{Nouri2023_Probabilistic}. 
A large historical dataset is used to compute lead-time-specific forecast errors, which are grouped according to atmospheric conditions represented by DNI variability classes to derive empirical error distributions. 
The resulting error quantiles are then applied to each deterministic forecast to obtain predictive quantiles.

The model was retrained using the same development dataset and data splits as the proposed method, following the training configuration described in the original publication~\cite{Fabel2024_Combining}.
This baseline serves as a representative state-of-the-art deep learning approach for direct solar forecasting~\cite{Nouri2025_Enhancing} and enables comparison between generative and end-to-end forecasting paradigms.

\subsubsection{SkyGPT}
SkyGPT~\cite{Nie2024_Sky} is a recently proposed transformer-based generative model for solar forecasting and is included as a state-of-the-art reference method. 
Similar to the decoupled baseline formulation considered in this work, SkyGPT follows a forecasting paradigm in which future sky images are first predicted and subsequently transformed into solar forecasts by a separate downstream model. 
In its original formulation, SkyGPT predicts photovoltaic (PV) power output rather than GHI. 
To this end, the generated future sky images are processed by a U-Net--based model, whose spatial output is aggregated to obtain a scalar forecast value. 
As for our CNN-based regressor, the U-Net model is trained separately on real images and applied during inference to synthetic images.

The video prediction component is based on VideoGPT and employs a vector-quantized variational autoencoder (VQ-VAE) to encode sky images into a discrete latent representation. 
Future latent tokens are then predicted autoregressively using a transformer backbone. 
The predicted latent states are subsequently decoded into future sky images, which serve as input to the downstream forecasting model.

For a fair comparison, all SkyGPT components were retrained using the same development dataset and data splits as the proposed method. 
The publicly available implementation provided by the authors~\cite{Nie2023_SkyGPT_GitHub} and the accompanying publication~\cite{Nie2024_SkyGPT} served as the basis for the implementation.
The original PV power prediction head was adapted to predict GHI instead, and minor modifications were required to ensure compatibility with the current software environment. 
Consequently, the reported results should be interpreted as a reimplementation under a common experimental setup rather than a direct reproduction of the results reported in the original publication.

\subsection{Evaluation}
\label{sec:eval}

All models were evaluated on the two test datasets described in Section~\ref{sec:dataset_construction}. 
To assess the impact of task-aware latent coupling, first representation learning with respect to image reconstruction and irradiance estimation is analyzed.
Forecasts of the presented model were generated for a horizon of 30 min at a temporal resolution of one minute. 
For probabilistic evaluation, an ensemble of $S=4$ stochastic trajectories was generated for each forecast initialization.
The resulting ensembles were converted into predictive distributions as described in Section~\ref{sec:uncertainty_quantification}.

The proposed approach is evaluated with respect to irradiance forecasting performance, ramp event prediction, and video prediction quality. 
All metrics are computed separately for each forecast lead time and subsequently averaged over the respective evaluation dataset.

\subsubsection{Deterministic Forecast Metrics}
\label{sec:det_metrics}

Deterministic forecast performance is assessed using the root-mean-square error (RMSE), mean absolute error (MAE), and mean bias error (MBE), which quantify overall forecast accuracy, average absolute deviations, and systematic forecast bias, respectively. In addition, forecast skill (FS) is computed relative to the Scaled Persistence baseline, enabling a normalized comparison of forecasting performance.

\begin{align}
\text{RMSE} &= \sqrt{\frac{1}{N}\sum_{i=1}^{N}\left(\hat{y}_i - y_i\right)^2}, \label{eq:rmse} \\
\text{MAE}  &= \frac{1}{N}\sum_{i=1}^{N}\left|\hat{y}_i - y_i\right|, \label{eq:mae} \\
\text{MBE}  &= \frac{1}{N}\sum_{i=1}^{N}\left(\hat{y}_i - y_i\right), \label{eq:mbe} \\
\text{FS}   &= 1 - \frac{\text{RMSE}_{\text{model}}}{\text{RMSE}_{\text{SP}}}. \label{eq:fs}
\end{align}

where $\hat{y}_i$ and $y_i$ denote forecasted and observed irradiance values, respectively. Positive forecast skill values indicate an improvement over the Scaled Persistence baseline.

\subsubsection{Probabilistic Forecast Metrics}
\label{sec:prob_metrics}

Since the proposed approach generates an ensemble of irradiance forecasts through stochastic latent-space sampling, forecast uncertainty can be quantified directly. Probabilistic forecast performance is evaluated using the continuous ranked probability score (CRPS), prediction interval coverage probability (PICP), and a probabilistic skill score based on CRPS (CRPSS). These metrics jointly assess forecast sharpness, reliability, and overall probabilistic forecast quality.

The PICP measures the fraction of observations falling within a prediction interval of nominal coverage $(1-\alpha)$:
\begin{equation}
\mathrm{PICP}_{\alpha}
= 
\frac{1}{N} \sum_{i=1}^{N} 
\mathbf{1}\!\left\{ y_i \in \left[L_{i,\alpha},\, U_{i,\alpha}\right] \right\},
\label{eq:picp}
\end{equation}

where $L_{i,\alpha}$ and $U_{i,\alpha}$ denote the lower and upper interval bounds. 
CRPSS is computed relative to a climatological reference forecast, CSD-Clim~\cite{Salle2021_new}:

\begin{align}
\text{CRPS} &= \frac{1}{N} \sum_{i=1}^{N} 
\int_{0}^{1} \left[ F_{\hat{y}_i}(x) - F_{y_i}(x) \right]^2 \, \mathrm{d}x,
\label{eq:crps} \\[6pt]
\text{CRPSs} &= 1 - \frac{\text{CRPS}_{\text{model}}}{\text{CRPS}_{\text{CSD-Clim}}}.
\label{eq:crpss}
\end{align}
with positive values indicating improved probabilistic performance over the reference model.

\subsubsection{Ramp Event Metrics}
\label{sec:ramp_metrics}

In addition to aggregate forecast errors, the ability to predict rapid irradiance fluctuations is assessed using a ramp event metric following~\cite{Nouri2024_Ramp}. 
Ramp events are defined as changes in the clear-sky index exceeding a threshold of $\varepsilon_r = 0.14$. 
To reflect that ramp forecasting primarily aims to predict the occurrence rather than the exact timing of irradiance ramps, a tolerance window of $\pm3\,\mathrm{min}$ is applied.

For stochastic forecasting models, ramp events are first detected independently for each generated trajectory. 
A ramp event is predicted at a given lead time if it is detected in at least 25\,\% of the generated trajectories. 
With four generated samples, this corresponds to detecting a ramp in at least one trajectory.
This aggregation strategy is applied consistently to all probabilistic forecasting models.

Based on the resulting event detections, true positives (TP), false positives (FP), false negatives (FN), and true negatives (TN) are computed. 
Standard classification metrics including precision, recall, and F1-score are then derived according to

\begin{align}
\mathrm{Precision}
&=
\frac{\mathrm{TP}}
{\mathrm{TP}+\mathrm{FP}},
\\
\mathrm{Recall}
&=
\frac{\mathrm{TP}}
{\mathrm{TP}+\mathrm{FN}},
\\
\mathrm{F1}
&=
\frac{
2 \cdot \mathrm{Precision}
\cdot \mathrm{Recall}
}
{
\mathrm{Precision}
+
\mathrm{Recall}
}.
\end{align}

\subsubsection{Video Prediction Metrics}

To assess the impact of the proposed latent-space coupling on the underlying video prediction task, image quality metrics are additionally evaluated on the generated future sky images. Besides irradiance forecasting performance, these metrics provide insight into whether the task-aware latent representation affects the ability of the model to represent cloud evolution.

Video prediction quality is assessed using pixel-wise MSE and MAE as well as the perceptual metrics structural similarity (SSIM) and peak signal-to-noise ratio (PSNR). Let $\hat{\mathrm{X}}_i$ and $\mathrm{X}_i$ denote predicted and reference images, respectively. 
To evaluate the quality of the predicted sky images, pixel-wise error metrics (MSE and MAE) as well as perceptual image quality metrics (SSIM and PSNR) are computed as

\begin{align}
\text{MSE}_{\text{img}} 
&= \frac{1}{N H W}
\sum_{i=1}^{N}
\sum_{h=1}^{H}
\sum_{w=1}^{W}
\left(
\hat{X}_{i}^{(h,w)} - X_{i}^{(h,w)}
\right)^2,
\label{eq:mse_img} \\[6pt]
\text{MAE}_{\text{img}} 
&= \frac{1}{N H W}
\sum_{i=1}^{N}
\sum_{h=1}^{H}
\sum_{w=1}^{W}
\left|
\hat{X}_{i}^{(h,w)} - X_{i}^{(h,w)}
\right|,
\label{eq:mae_img} \\[6pt]
\text{SSIM} 
&= \frac{1}{N}
\sum_{i=1}^{N}
\frac{
\left(2 \mu_{\hat{X}_{i}} \mu_{X_{i}} + C_1\right)
\left(2 \sigma_{\hat{X}_{i} X_{i}} + C_2\right)
}{
\left(\mu_{\hat{X}_{i}}^2 + \mu_{X_{i}}^2 + C_1\right)
\left(\sigma_{\hat{X}_{i}}^2 + \sigma_{X_{i}}^2 + C_2\right)
},
\label{eq:ssim} \\[6pt]
\text{PSNR} 
&= 10 \log_{10}
\left(
\frac{I_{\max}^2}{\text{MSE}_{\text{img}}}
\right),
\label{eq:psnr} \\
\end{align}

where $N$ denotes the number of evaluated image pairs, $H$ and $W$ are the image height and width, respectively, $X_i^{(h,w)}$ and $\hat{X}_i^{(h,w)}$ denote the ground-truth and predicted pixel intensities at pixel location $(h,w)$ of the $i$-th image, $\mu$, $\sigma^2$, and $\sigma_{\hat{X}_iX_i}$ denote the local mean, variance, and covariance used in the SSIM computation, $C_1$ and $C_2$ are stability constants, and $I_{\max}$ is the maximum possible pixel intensity.
All image metrics are evaluated separately for each forecast lead time.

\subsubsection{Evaluation by Atmospheric Conditions}
\label{sec:dni_var_classes}

In addition to the aggregated evaluation over the complete test datasets, forecasting performance is analyzed separately for different atmospheric conditions. 
For this purpose, all forecast samples are grouped according to the DNI variability classification proposed by~\cite{SchroedterHomscheidt2018_Classifying}, which combines the magnitude and short-term variability of DNI into eight different classes. 
Since the resulting variability classes have been shown to correlate well with prevailing cloud conditions, they provide a meaningful basis for analyzing forecasting performance in different meteorological situations.

Following~\cite{Nouri2025_Enhancing}, the eight DNI variability classes are aggregated into three broader categories: \emph{mostly clear-sky} (classes~1--2), \emph{scattered clouds} (classes~3--6), and \emph{mostly overcast} (classes~7--8). 
By reporting evaluation metrics also separately for these three atmospheric categories, the robustness, strengths and weaknesses of the different forecasting approaches can be assessed under varying cloud conditions.

\section{Results \& Discussion}
\label{sec:results}
This section first investigates the effect of the proposed latent-space coupling mechanism before comparing the complete FarSky framework with state-of-the-art forecasting approaches.

\subsection{Evaluation of Latent-Space Coupling}
The first part of the evaluation aims to isolate the effect of task-aware latent-space coupling from the remaining components of the forecasting pipeline. 
To this end, the proposed DCAEIrr and FarSky models are compared with their corresponding decoupled counterparts. 
The analysis first investigates the learned latent representations in terms of irradiance estimation and image reconstruction before evaluating their impact on deterministic, probabilistic, and ramp event forecasting performance.
Evaluation is conducted on the \PSABench{} dataset throughout this section.

\subsubsection{Task-Aware Latent Representation Learning}
\label{sec:results_rep_learn}
Before evaluating the forecasting performance of the proposed framework, the DCAEIrr is analyzed independently to assess whether latent-space coupling through a multi-task learning objective leads to more informative latent representations. 
To this end, the DCAEIrr is evaluated on real sky images and compared to the standalone irradiance regression model (CNN) used in the decoupled baseline described in Section~\ref{sec:decoupled_baseline}. 
In particular, the irradiance estimates obtained from the DCAEIrr are compared to those of the standalone CNN to assess the benefit of jointly learning image reconstruction and irradiance estimation. 
In addition, the image reconstruction quality of the DCAEIrr is compared to that of the original DCAE to quantify the effect of incorporating irradiance supervision into the latent representation.

\begin{figure}[htbp]
    \centering
    \includegraphics[width=0.8\linewidth]{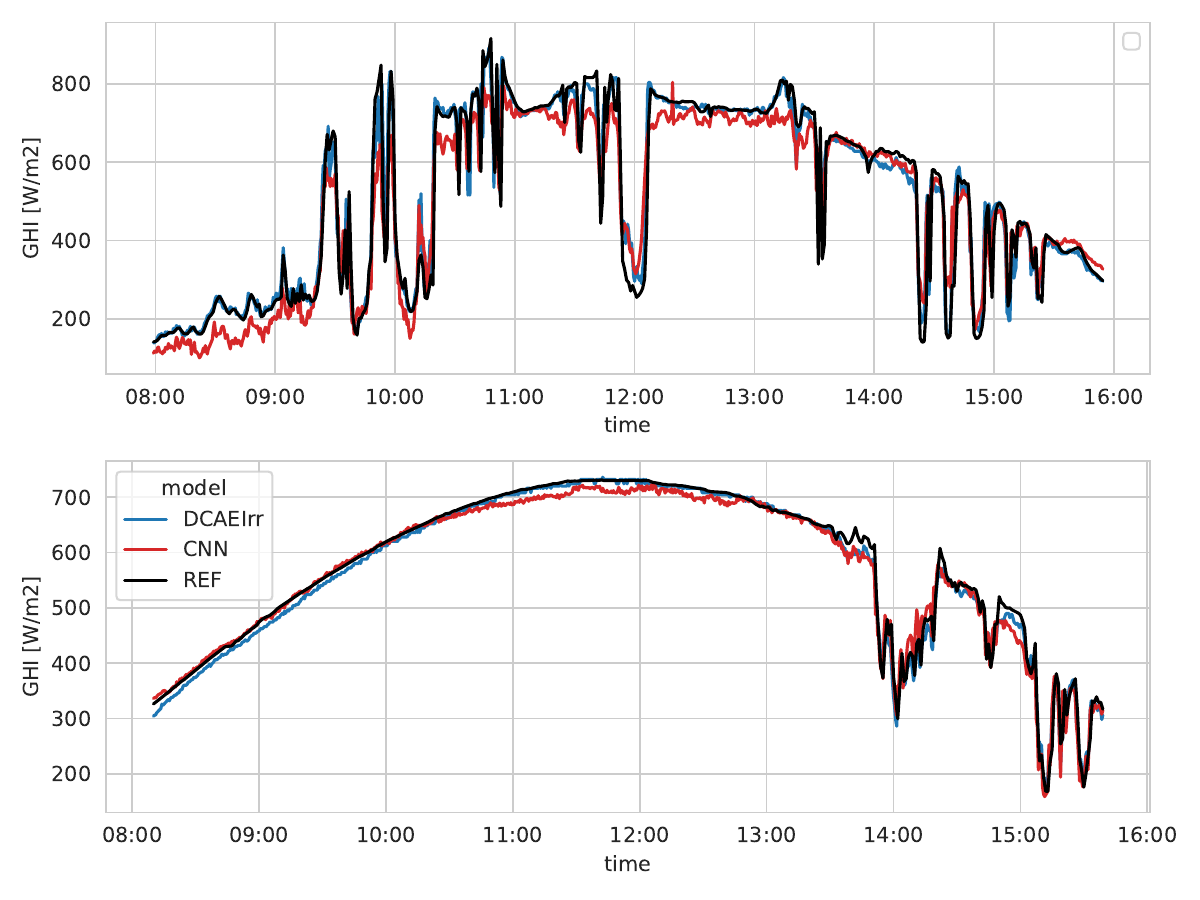}
    \caption{Exemplary GHI predictions from real all-sky images (lead time zero) obtained from the enhanced autoencoder DCAEIrr and a standalone CNN-based irradiance regressor for two days from the \PSABench{} dataset.}
    \label{fig:replearn_irr_pred_examples}
\end{figure}

Figure~\ref{fig:replearn_irr_pred_examples} shows measured and predicted GHI for two representative days from the benchmark dataset.
Both models reproduce the overall irradiance evolution well under clear-sky and cloudy conditions, indicating that relevant cloud features can be extracted from individual all-sky images. 
The largest deviations occur during periods of highly variable irradiance, where the relationship between cloud appearance and surface irradiance becomes particularly complex. 
In these situations, the DCAEIrr generally follows the reference measurements more closely than the standalone CNN. 
Furthermore, the CNN exhibits a systematic tendency to underestimate irradiance, whereas the DCAEIrr remains more closely aligned with the measured values.

\begin{table}[htbp]
    \centering
    \caption{Irradiance estimation errors for the irradiance regressor (CNN) evaluated on real sky images at lead time zero compared to the task-aware DCAEIrr on the \PSABench{} dataset.}
    \begin{tabular}{lccc}
    \toprule
    Input & RMSE $\downarrow$ [W\,m$^{-2}$] & MAE $\downarrow$ [W\,m$^{-2}$] & MBE [W\,m$^{-2}$] \\
    \midrule
    CNN   & 52.0 & 37.5 & -24.9 \\
    DCAEIrr & \textbf{25.9} & \textbf{14.8} & \textbf{-5.7} \\
    \bottomrule
    \end{tabular}
    \label{tab:error_metrics_dcaeirr}
\end{table}

Quantitative results are summarized in Table~\ref{tab:error_metrics_dcaeirr}. 
Compared to the CNN, the DCAEIrr substantially improves irradiance estimation performance on unseen test data. 
RMSE decreases from 52.0 to 25.9\,W\,m$^{-2}$, while MAE is reduced from 37.5 to 14.8\,W\,m$^{-2}$. 
The largest improvement is observed for the mean bias error, whose magnitude decreases by almost 80\,\%, from $-24.9$ to $-5.7$\,W\,m$^{-2}$. The consistently lower error metrics indicate that the multi-task learning objective encourages task-aware latent representations that encode irradiance-relevant information more effectively than a model trained solely for irradiance regression. 
Furthermore, the substantial reduction in systematic bias suggests improved robustness when applied to previously unseen data.

\begin{table}[htbp]
\centering
\caption{Image reconstruction error metrics and scores comparing DCAE and DCAEIrr on the \PSABench{} dataset. Errors represent average pixel-wise differences for the range [0,~255].}
\begin{tabular}{lcccc}
\toprule
Model   & MAE           & MSE           & PSNR          & SSIM          \\
\midrule
DCAE    & \textbf{0.86} & \textbf{4.38} & \textbf{45.6} & \textbf{0.99} \\
DCAEIrr & 1.09          & 6.07          & 43.4          & 0.98          \\
\bottomrule
\end{tabular}
\label{tab:image_error_metrics_dcaeirr}
\end{table}

To assess whether these improvements come at the expense of visual fidelity, the image reconstruction performance of the DCAEIrr is compared to the original DCAE in Table~\ref{tab:image_error_metrics_dcaeirr}. 
Incorporating the additional irradiance objective introduces a small degradation in reconstruction quality, reflected by slightly higher pixel-wise reconstruction errors and marginally lower PSNR and SSIM values. 
Nevertheless, the differences remain small, with the DCAEIrr still achieving reconstruction performance close to that of the original DCAE. 
This behavior is consistent with the intended design of the model, where part of the latent capacity is reallocated from purely visual information towards irradiance-relevant features.

Taken together, these results demonstrate that latent-space coupling successfully embeds irradiance information into the latent representations while largely preserving image reconstruction quality. 
The resulting task-aware latent space provides a substantially improved basis for irradiance estimation and forms the foundation for the forecasting framework evaluated in the following sections.

\subsubsection{Latent-Space Video Prediction}

Having demonstrated that latent-space coupling produces more informative representations for irradiance estimation, the next question is how these task-aware latent representations affect generative video prediction performance.
To investigate this aspect, the original FAR model is compared to FarSky, which employs the same diffusion-based forecasting architecture but operates on the irradiance-aware latent space learned by the DCAEIrr.

\begin{figure}[htbp]
\centering
\includegraphics[width=\textwidth]{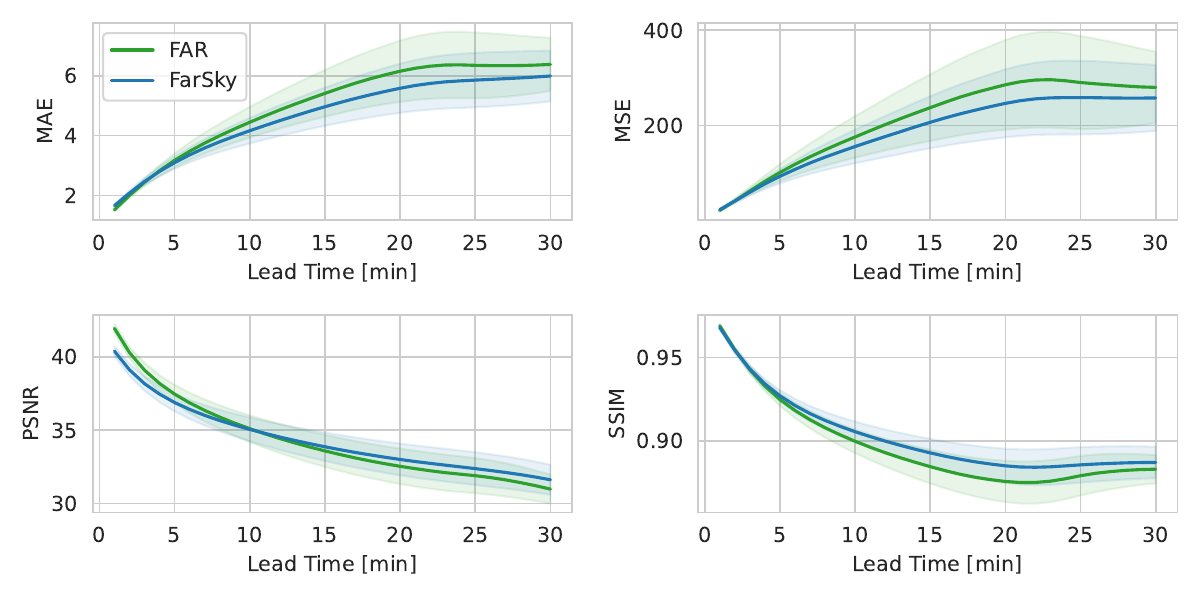}
\caption{Comparison of image metrics over lead lead time for video prediction models FAR and FarSky on the \PSABench{} dataset. Shaded areas indicate the mean standard deviation across multiple generated image samples for each lead time, reflecting variability in the stochastic predictions.}
\label{fig:farsky_image_metrics}
\end{figure}

Figure~\ref{fig:farsky_image_metrics} summarizes the lead time-dependent image quality metrics obtained on the benchmark dataset. 
At very short lead times, the original FAR model achieves marginally better agreement with the reference images, most notably in terms of PSNR. 
This behavior is consistent with the slightly superior reconstruction performance of the original DCAE observed in the previous section and can therefore be attributed to the absence of the additional irradiance supervision during representation learning.

For increasing lead times, however, FarSky consistently achieves slightly improved image quality across all evaluated metrics. 
Although the absolute differences remain small, the advantage persists throughout most of the forecast horizon.
This result suggests that the irradiance-aware latent representations provide additional constraints on the latent dynamics learned by the diffusion model, leading to more accurate predictions of future sky states.

Overall, the results indicate that incorporating irradiance supervision during representation learning does not compromise video prediction performance. 
On the contrary, the task-aware latent space enables a modest but consistent improvement in forecasting quality while simultaneously providing direct access to irradiance information. 
These findings suggest that the benefits of latent-space coupling extend beyond irradiance estimation and also support the prediction of future cloud evolution.

\subsubsection{Deterministic Irradiance Forecasting}

\begin{figure}[t]
    \centering
    \includegraphics[width=\linewidth]{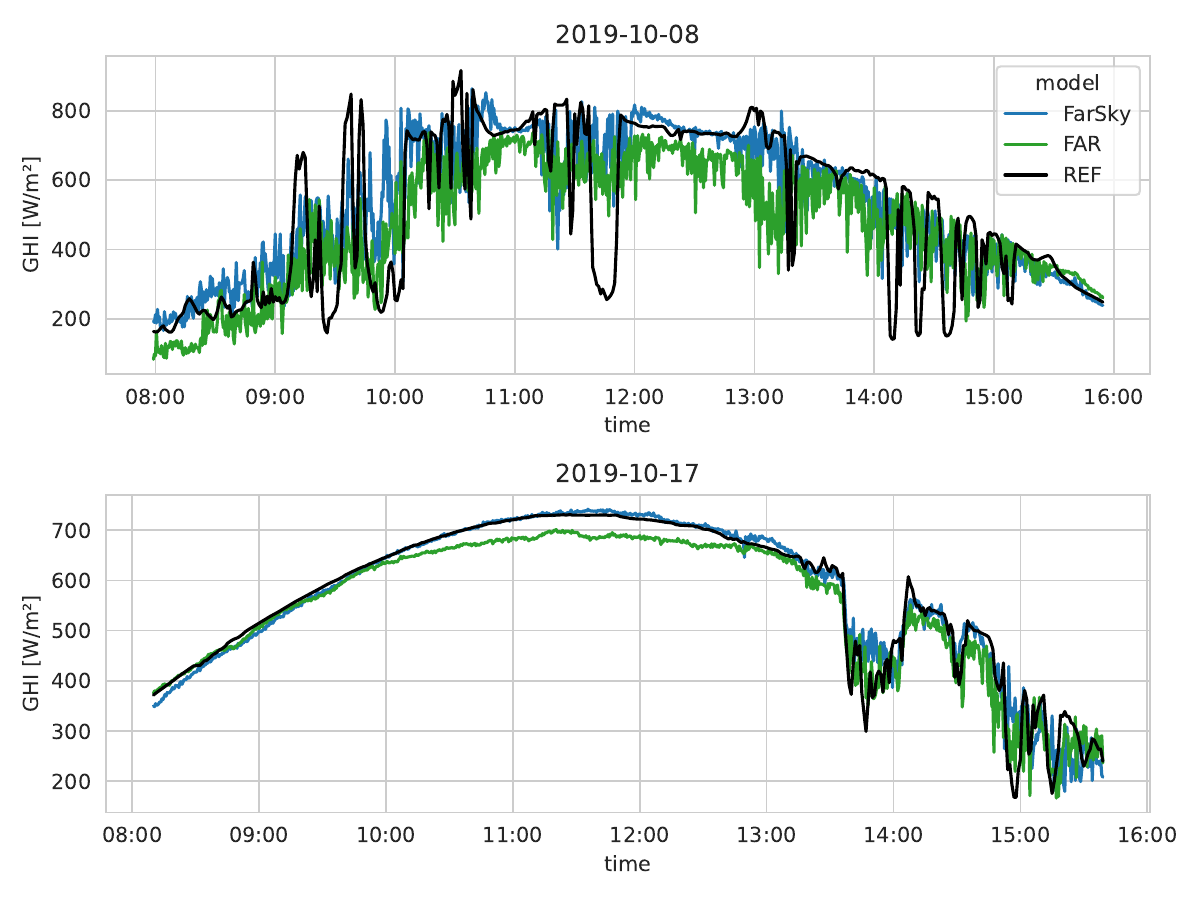}
    \caption{Exemplary deterministic GHI forecasts at lead time 15\,min obtained from the proposed FarSky framework and the decoupled FAR baseline for two days from the \PSABench{} dataset.}
    \label{fig:farsky_forecast_examples}
\end{figure}

Before comparing the aggregated forecasting metrics, Figure~\ref{fig:farsky_forecast_examples} illustrates mean GHI forecasts at a lead time of 15\,min obtained from the proposed FarSky framework and the decoupled FAR baseline for the same exemplary days presented in the previous section. 
For the day with pronounced cloud-induced variability (top), both approaches capture the overall irradiance evolution, while FarSky more closely follows the measured GHI and exhibits a substantially smaller systematic underestimation than the decoupled FAR pipeline. 
Neither approach fully reproduces the magnitude of the observed short-term fluctuations, which is partly attributable to averaging the stochastic forecast trajectories. 
During the predominantly clear-sky period shown in the second example (bottom), both methods accurately reproduce the overall irradiance evolution. 
However, the decoupled FAR pipeline again exhibits a pronounced negative bias around solar noon, whereas the FarSky forecast remains in much closer agreement with the measurements.

\begin{figure}[htbp]
\centering
\includegraphics[width=\textwidth]{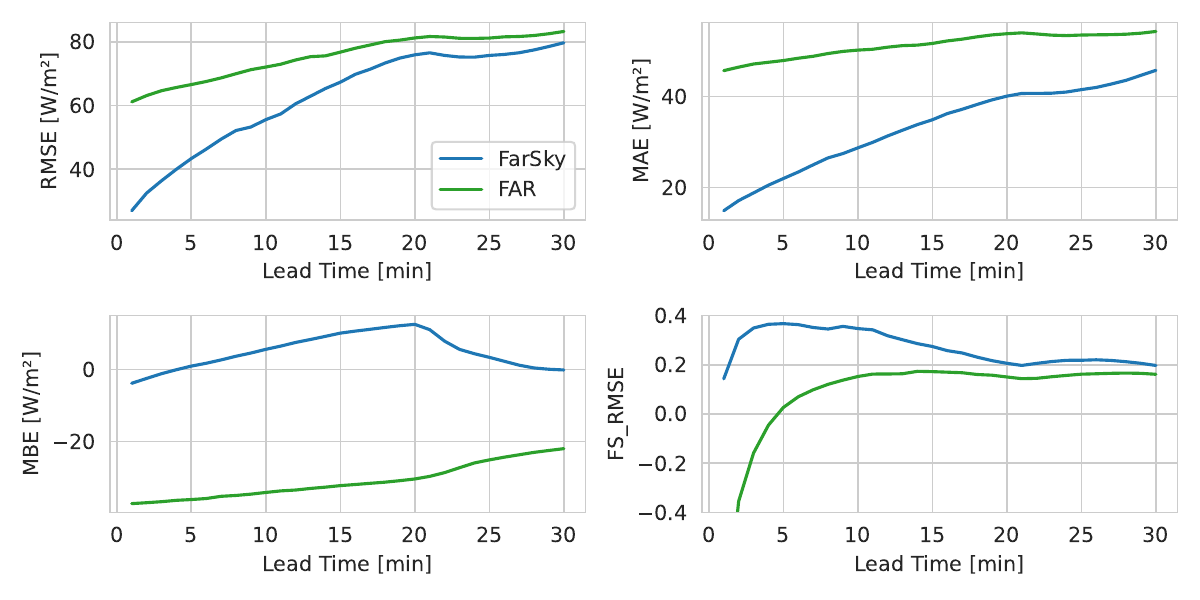}
\caption{Comparison of deterministic error metrics over lead time for irradiance  forecasts (GHI) obtained from FAR combined with the irradiance regressor (CNN) and from FarSky. Shown are RMSE, MAE, MBE, and forecast skill (FS) based on RMSE, computed relative to persistence forecasts. All metrics are evaluated using the ensemble mean forecast.}
\label{fig:farsky_det_metrics}
\end{figure}

Deterministic forecasting performance of the proposed FarSky framework is evaluated using the metrics described in Section~\ref{sec:det_metrics}. 
Figure~\ref{fig:farsky_det_metrics} compares the lead time-dependent performance of FarSky with the decoupled baseline, where irradiance is estimated from generated sky images using the standalone CNN regressor.

Across all lead times, FarSky consistently achieves lower RMSE and MAE than the decoupled approach. 
The largest improvements are observed at short forecast horizons, where the error metrics remain only slightly above the values obtained from direct irradiance decoding of real images in Section~\ref{sec:results_rep_learn}. 
This indicates that the latent representations predicted by the generative model remain close to those obtained from real observations, enabling accurate irradiance forecasts immediately beyond the forecasting origin.

The most pronounced differences between both approaches are observed for the bias. 
At the first forecast lead time of one minute, the decoupled FAR pipeline exhibits an MBE of approximately $-38$\,W\,m$^{-2}$, indicating a substantially stronger negative bias than the bias of the standalone CNN evaluated on real images ($-24.9$\,W\,m$^{-2}$). 
This additional bias indicates that the reconstruction and forecasting steps introduce a domain shift between generated and real images, which propagates through the downstream irradiance regressor. 
In contrast, FarSky maintains a near-zero bias at short lead times, with an MBE of $-3.8$\,W\,m$^{-2}$ compared to $-5.7$\,W\,m$^{-2}$ for direct irradiance decoding from real images. 
This observation supports the hypothesis that latent-space coupling effectively mitigates interface-level domain shift effects by avoiding the intermediate image-to-irradiance mapping required by the decoupled pipeline.

The reduction in systematic bias directly translates into improved overall forecast accuracy and forecast skill. 
FarSky achieves forecast skill values exceeding 30\,\% on the \PSABench{} dataset, substantially outperforming the decoupled baseline across the entire forecasting horizon. 
Although errors increase with lead time for both approaches, the relative advantage of FarSky remains largely preserved.

An additional observation is the evolution of the forecast bias with lead time.
During the first 20 minutes, the bias of FarSky gradually increases from values close to zero to a moderate overestimation of irradiance, suggesting that the generated future sky states exhibit, on average, less attenuation of solar irradiance than observed.
Beyond approximately 20 minutes, however, the bias decreases again and approaches zero. The origin of this behavior is currently unclear, but it may be related to the fact that inference is performed autoregressively beyond the maximum prediction horizon encountered during training. 
More generally, the reduction in bias at longer lead times may reflect a tendency of the generative model to converge toward climatologically more typical sky states as forecast uncertainty increases.
Nevertheless, the substantially lower MAE achieved by FarSky throughout the forecasting horizon demonstrates that irradiance-aware latent representations provide a more effective basis for solar forecasting than conventional approaches based either on generated images and a separate irradiance regressor.

Overall, these results demonstrate that latent-space coupling not only improves irradiance estimation from real sky images but also translates into substantially improved deterministic forecasting performance. 
The benefits are particularly pronounced with respect to systematic bias, highlighting the importance of jointly learning forecasting and irradiance-relevant representations within a shared latent space.

\subsubsection{Probabilistic Irradiance Forecasting}

\begin{figure}[htbp]
\centering
\includegraphics[width=\textwidth]{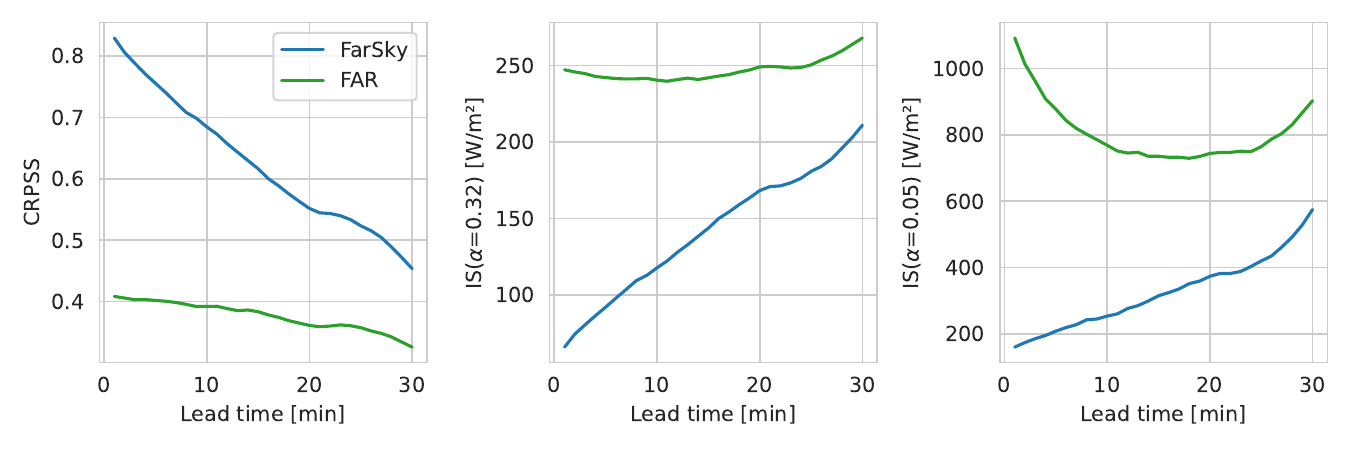}
\caption{Comparison of probabilistic error metrics over lead time for irradiance  forecasts (GHI) obtained from FAR combined with the irradiance regressor (CNN) and from FarSky. Shown are CRPSS and IS for the 68\% and 95\% prediction intervals.}
\label{fig:farsky_prob_metrics}
\end{figure}

Beyond deterministic accuracy, generative forecasting approaches provide the additional capability of estimating forecast uncertainty through multiple stochastic prediction trajectories.
Figure~\ref{fig:farsky_prob_metrics} compares the probabilistic forecasting performance of FarSky and the decoupled FAR pipeline using the CRPSS and IS for the 68\,\% and 95\,\% prediction intervals, calculated for each lead time.

FarSky consistently achieves higher CRPSS values than the decoupled approach across all lead times. 
The largest improvements are observed at short forecast horizons, where the probabilistic skill is nearly twice as high as that of the decoupled pipeline. 
Although CRPSS gradually decreases with increasing lead time for both approaches, reflecting the growing uncertainty associated with longer forecasting horizons, the advantage of FarSky is maintained throughout the entire forecast horizon. 
These results indicate that latent-space coupling not only improves deterministic point forecasts but also yields more accurate probabilistic predictions.

The interval scores provide a consistent picture. For both the 68\,\% and 95\,\% prediction intervals, FarSky achieves substantially lower scores than the decoupled pipeline across all lead times, indicating a better trade-off between interval sharpness and coverage. 
While the interval scores generally increase with forecast horizon, reflecting growing uncertainty, the gap between the two methods decreases at longer lead times. 
Notably, the interval scores of the decoupled FAR pipeline do not exhibit a strictly monotonic increase, which suggests inconsistencies in the calibration of its predictive distributions.

Together with the deterministic results, these findings suggest that the systematic negative bias of the decoupled pipeline propagates to the predictive distributions, shifting them toward lower irradiance values.
Consequently, observations are more likely to exceed the upper prediction bounds, which contributes to inferior interval scores and indicates a tendency toward undercoverage compared to FarSky.

\subsubsection{Ramp Event Prediction}

Finally, the impact of latent-space coupling on ramp event prediction is evaluated. 
Unlike the deterministic and probabilistic metrics discussed previously, ramp detection is based on relative irradiance changes rather than absolute irradiance values. 
Consequently, systematic forecast bias is expected to have a smaller influence on classification performance.

\begin{figure}[htbp]
\centering
\includegraphics[width=\textwidth]{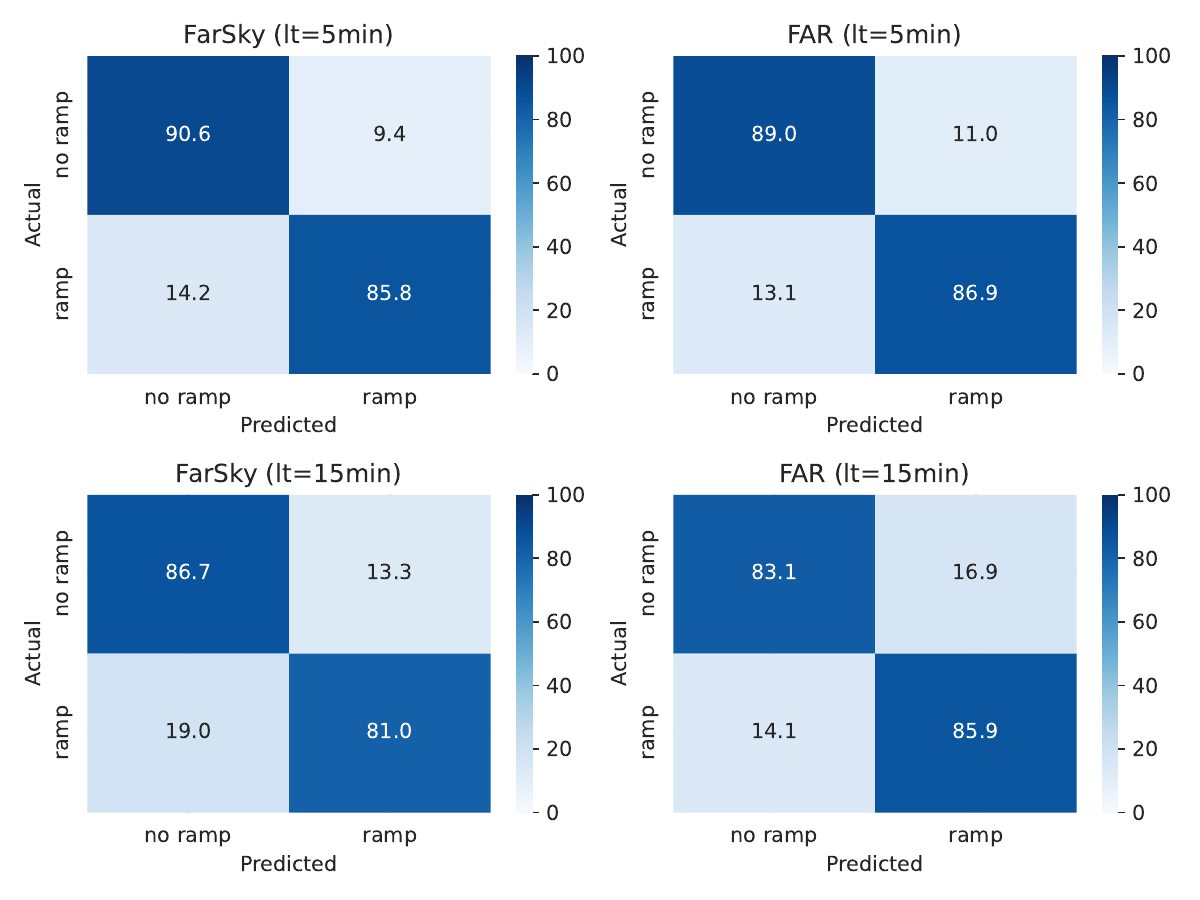}
\caption{Confusion matrices for ramp detection at forecast lead times 5 and 15 minutes, comparing FarSky and the decoupled baseline. Rows correspond to the observed classes and are normalized such that each row sums to 100\%.}
\label{fig:cm_5_15}
\end{figure}

Figure~\ref{fig:cm_5_15} compares the normalized confusion matrices of FarSky and the decoupled FAR pipeline for representative forecast lead times of 5 and 15\,min.
At the shorter forecast horizon, both approaches exhibit very similar classification performance, correctly identifying approximately 86\,\% of ramp events while maintaining true negative rates close to 90\,\%. 
Differences between the two methods remain minor, with FarSky exhibiting a slightly lower false-positive rate at the expense of a marginal reduction in true-positive detections.

At a lead time of 15\,min, this trend becomes more pronounced. 
FarSky correctly identifies a larger fraction of non-ramp situations than the decoupled pipeline (86.7\,\% versus 83.1\,\%), whereas FAR detects a slightly larger proportion of actual ramp events (85.9\,\% versus 81.0\,\%). 
This indicates that latent-space coupling leads to a slightly more conservative classification behavior, reducing false alarms while accepting a modest increase in missed ramp events. 
Overall, however, the differences remain comparatively small.

Unlike the deterministic and probabilistic forecasting results, latent-space coupling therefore provides no substantial improvement for ramp event prediction. 
This observation supports the interpretation established in the previous subsections. 
The primary benefit of the proposed approach lies in reducing systematic irradiance bias rather than improving the prediction of relative irradiance variability. 
Since ramp events are defined by changes in irradiance over time, they are inherently less sensitive to constant offsets in the forecast than the deterministic error metrics.

\subsection{Benchmark Evaluation}
After analyzing the effect of latent-space coupling in isolation, the complete FarSky framework is compared against the benchmark models introduced in Section~\ref{sec:comparison_methods}.
The evaluation is conducted on both test datasets and considers deterministic accuracy, probabilistic forecast quality, reliability, and ramp event prediction.
Unless stated otherwise, all benchmark metrics are averaged over the common forecast horizon of the first 15\,min.

\subsubsection{Overall Forecasting Performance}

\begin{table}[t]
\centering
\caption{Deterministic forecasting performance averaged over the first 15\,min forecast horizon. Lower values indicate better performance for RMSE and MAE, while higher values indicate better performance for forecast skill (FS).}
\begin{tabular}{llcccc}
\toprule
Dataset & Model & RMSE [W\,m$^{-2}$] $\downarrow$ & MAE [W\,m$^{-2}$] $\downarrow$ & MBE [W\,m$^{-2}$] & FS $\uparrow$ \\
\midrule
\multirow{4}{*}{\PSABench{}}
& FarSky  & \textbf{50.0} & \textbf{25.8} & 3.6  & \textbf{0.320} \\
& FAR     & 69.7 & 49.1 & -34.9 & -0.005 \\
& E2E-TF  & 57.7 & 32.1 & -4.0 & 0.209 \\
& SkyGPT  & 86.8 & 63.4 & \textbf{1.3 }& -0.292 \\
\midrule
\multirow{4}{*}{\SVATest{}}
& FarSky  & \textbf{98.0} & \textbf{50.6} & 15.8 & \textbf{0.241} \\
& FAR     & 122.8 & 80.2 & -44.4 & 0.018 \\
& E2E-TF  & 100.8 & 55.3 & \textbf{-7.6} & 0.209 \\
& SkyGPT  & 146.7 & 108.2 & -20.9 & -0.214 \\
\bottomrule
\end{tabular}
\label{tab:deterministic_benchmark_comparison}
\end{table}

Table~\ref{tab:deterministic_benchmark_comparison} summarizes the deterministic forecasting performance.
On the \PSABench{} dataset, FarSky achieves the best performance across the main deterministic metrics, reducing RMSE from 57.7\,W\,m$^{-2}$ for E2E-TF and 69.7\,W\,m$^{-2}$ for FAR to 50.0\,W\,m$^{-2}$.
This corresponds to the highest forecast skill of 32.0\,\%. 
The performance gap is even more pronounced relative to the SkyGPT-based baseline, which does not outperform persistence under the considered evaluation setup.
The same ranking is observed on the independent \SVATest{} dataset. 
FarSky again achieves the lowest RMSE and MAE as well as the highest forecast skill, improving RMSE from 100.8 to 98.0\,W\,m$^{-2}$ and increasing forecast skill from 20.9\,\% to 24.1\,\% compared to E2E-TF. 
Interestingly, all models exhibit substantially higher absolute error metrics and lower forecast skill on this dataset, highlighting the strong dependence of reported performance on the underlying data characteristics. 
Despite the smaller performance gap compared to the \PSABench{} dataset, these results demonstrate that the generative FarSky approach consistently outperforms an end-to-end model explicitly optimized for MSE, demonstrating the effectiveness of the proposed framework for accurate irradiance forecasting over short forecast horizons of up to 15\,min.

\begin{table}[t]
\centering
\caption{Probabilistic forecasting performance averaged over the first 15\,min forecast horizon. Higher values indicate better performance for CRPSS, while lower values indicate better performance for the interval score (IS).}
\begin{tabular}{llccc}
\toprule
Dataset & Model & CRPSS $\uparrow$ & IS$_{68}$ [W\,m$^{-2}$] $\downarrow$ & IS$_{95}$ [W\,m$^{-2}$] $\downarrow$ \\
\midrule
\multirow{4}{*}{\PSABench{}}
& FarSky & \textbf{0.715} & \textbf{107.0} & \textbf{236.5} \\
& FAR    & 0.396 & 242.3 & 838.8 \\
& E2E-TF & 0.677 & 130.3 & 266.6 \\
& SkyGPT & 0.275 & 276.9 & 636.3 \\
\midrule
\multirow{4}{*}{\SVATest{}}
& FarSky & \textbf{0.540} & \textbf{217.0} & 572.0 \\
& FAR    & 0.213 & 385.0 & 1308.5 \\
& E2E-TF & 0.532 & 233.6 & \textbf{517.2} \\
& SkyGPT & -0.070 & 523.9 & 1626.0 \\
\bottomrule
\end{tabular}
\label{tab:probabilistic_benchmark_comparison}
\end{table}

Table~\ref{tab:probabilistic_benchmark_comparison} summarizes the probabilistic forecasting performance. 
Across both datasets, FarSky achieves the highest CRPSS, indicating the best overall probabilistic forecast skill. 
On the \PSABench{} dataset, FarSky attains a CRPSS of 0.715, outperforming E2E-TF (0.677), the decoupled FAR pipeline (0.396), and SkyGPT (0.275).
A similar ranking is obtained on the \SVATest{} dataset, where FarSky again achieves the highest CRPSS (0.540), closely followed by E2E-TF (0.532).

The interval scores provide further insight into the quality of the prediction intervals. 
FarSky consistently achieves the lowest interval score for the 68\,\% prediction interval on both datasets, indicating the best trade-off between sharpness and coverage for the central part of the predictive distribution. 
For the wider 95\,\% prediction interval, FarSky achieves the best performance on the \PSABench{} dataset, while E2E-TF performs slightly better on the \SVATest{} dataset. 
This difference suggests that the comparatively poor reliability of the generative approaches can limit their performance for wider intervals, and may explain why FarSky does not consistently outperform E2E-TF across all settings. 
Nevertheless, despite deriving predictive distributions directly from stochastic generative trajectories without explicit calibration, FarSky achieves the highest overall probabilistic forecast skill on both datasets.

\begin{figure}[t]
\centering

\begin{subfigure}[t]{0.48\textwidth}
    \centering
    \includegraphics[width=\linewidth]{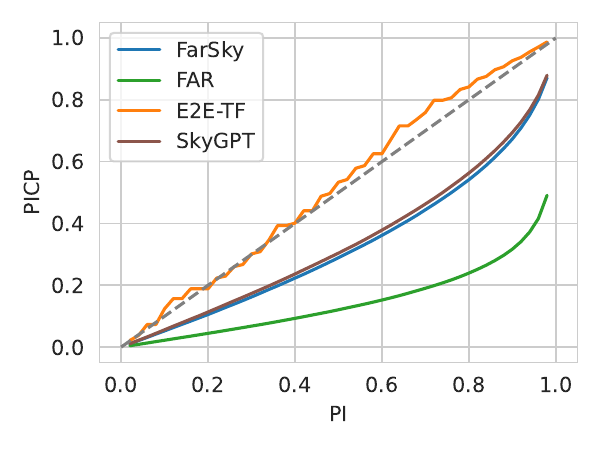}
    \caption{\PSABench{}}
    \label{fig:picp_pi_psa}
\end{subfigure}
\hfill
\begin{subfigure}[t]{0.48\textwidth}
    \centering
    \includegraphics[width=\linewidth]{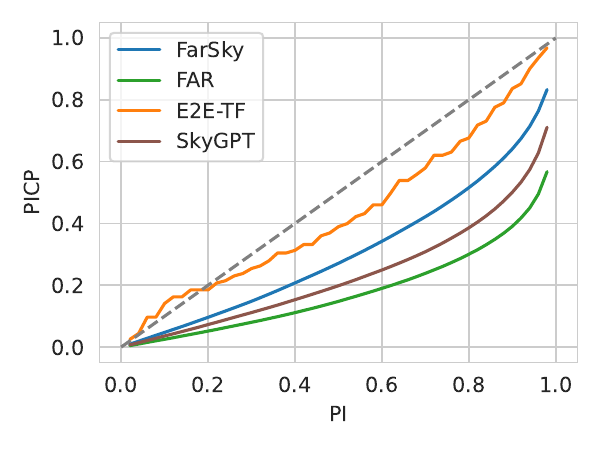}
    \caption{\SVATest{}}
    \label{fig:picp_pi_sva}
\end{subfigure}
\caption{Reliability diagrams comparing empirical prediction interval coverage probability (PICP) with nominal prediction interval (PI) for the evaluated probabilistic forecasting models. Results are averaged over the first 15\,min forecast horizon. The diagonal indicates perfect calibration.}
\label{fig:picp_pi}
\end{figure}

Figure~\ref{fig:picp_pi} compares the empirical coverage probability (PICP) with the nominal prediction interval (PI) for both test datasets. 
All generative models exhibit a certain degree of undercoverage, indicating that observations fall outside the prediction intervals more frequently than expected.
Among the evaluated models, E2E-TF achieves the closest agreement with the ideal calibration curve.

A plausible explanation for the remaining undercoverage of the generative approaches is the limited number of stochastic forecast samples used to estimate the predictive distributions. 
In this work, only four trajectories are generated per forecast due to the computational cost of the diffusion model, and the resulting samples are subsequently approximated by a Student's-$t$ distribution. 
Such a small sample size is unlikely to fully capture the variability represented by the generative model and may therefore lead to prediction intervals that are too narrow. 
Increasing the number of generated trajectories would likely provide a more faithful approximation of the predictive distribution and could further improve forecast calibration.

The strong calibration of the E2E-TF model is likely attributable to its probabilistic post-processing framework, which constructs prediction intervals from empirical historical forecast error distributions rather than estimating uncertainty directly from the forecasting model. 
Although this approach achieves better calibration, it requires an additional post-processing stage and dataset-specific error statistics. 
In contrast, FarSky derives probabilistic forecasts directly from stochastic generative trajectories without explicit calibration based on historical forecast errors.

\begin{table}[t]
\centering
\caption{Ramp event prediction performance averaged over the first 15\,min forecast horizon. Higher values indicate better performance for all metrics.}
\begin{tabular}{llcccc}
\toprule
Dataset & Model & Accuracy $\uparrow$ & Precision $\uparrow$ & Recall $\uparrow$ & F1 $\uparrow$ \\
\midrule
\multirow{4}{*}{\PSABench{}}
& FarSky & 0.887 & \textbf{0.490} & 0.839 & \textbf{0.618} \\
& FAR    & 0.869 & 0.452 & \textbf{0.862} & 0.592 \\
& E2E-TF & \textbf{0.891} & 0.340 & 0.002 & 0.004 \\
& SkyGPT & 0.707 & 0.298 & 0.542 & 0.384 \\
\midrule
\multirow{4}{*}{\SVATest{}}
& FarSky & \textbf{0.838} & 0.578 & 0.805 & 0.673 \\
& FAR    & 0.830 & 0.561 & \textbf{0.859} & \textbf{0.678} \\
& E2E-TF & 0.794 & \textbf{0.757} & 0.004 & 0.021 \\
& SkyGPT & 0.649 & 0.372 & 0.502 & 0.427 \\
\bottomrule
\end{tabular}
\label{tab:ramp_benchmark_comparison}
\end{table}

Table~\ref{tab:ramp_benchmark_comparison} reports the ramp event prediction performance. 
FarSky and the decoupled FAR pipeline achieve similar F1 scores on both datasets, confirming that latent-space coupling has only a limited effect on ramp detection. 
On the \PSABench{} dataset, FarSky achieves a slightly higher F1 score than FAR, whereas FAR performs marginally better on the \SVATest{} dataset. 
This behavior is consistent with the previous analysis: latent-space coupling primarily reduces systematic irradiance bias, while ramp events are defined by relative irradiance changes and are therefore less sensitive to absolute forecast offsets.

The E2E-TF baseline illustrates why accuracy alone is not sufficient for evaluating ramp event prediction. 
Although E2E-TF achieves high accuracy, its recall is close to zero on both datasets, resulting in very low F1 scores close to zero as well. 
This confirms the previously discussed problem of MSE optimization on forecast error, leading to smooth forecast curves that rarely represents ramp events. 
SkyGPT, as a generative model, performs substantially better than E2E-TF but still shows noticeably lower ramp detection performance than the diffusion-based approaches, particularly in terms of F1 score.

The benchmark comparison demonstrates that FarSky consistently achieves competitive or superior performance across the evaluated deterministic, probabilistic, and ramp forecasting metrics.
To further analyze the strengths and limitations of the different forecasting approaches, the following subsection examines their performance under different atmospheric conditions.

\subsubsection{Performance under Different Atmospheric Conditions}

\begin{figure}[htbp]
\centering
\includegraphics[width=\textwidth]{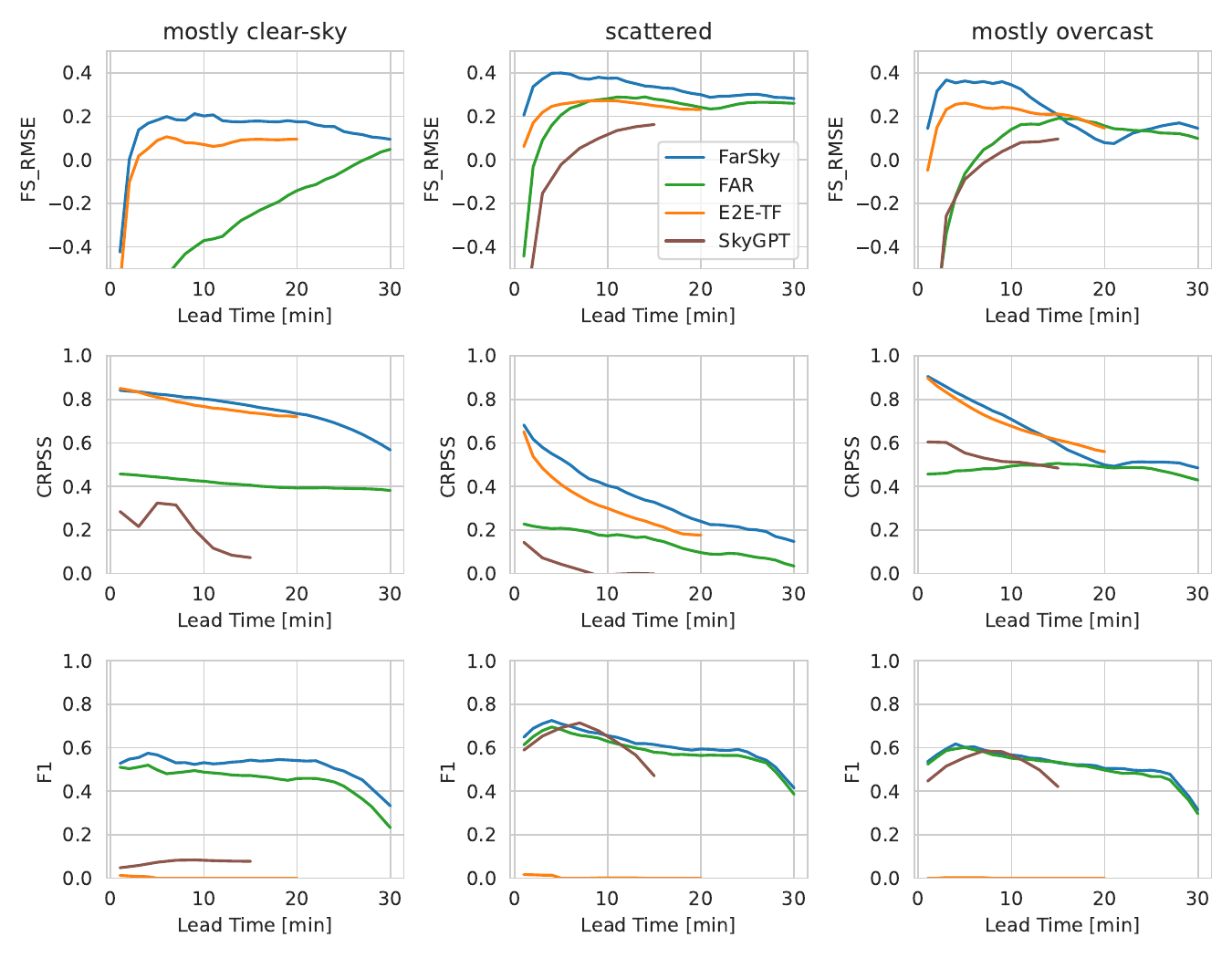}
\caption{FS, CRPSS, and ramp detection F1 score of the benchmark models as a function of forecast lead time for different atmospheric conditions on the \PSABench{} dataset. Columns correspond to mostly clear-sky, scattered-cloud, and mostly overcast conditions. Negative metric values are truncated to improve readability.}
\label{fig:metrics_psa_classes}
\end{figure}

\begin{figure}[htbp]
\centering
\includegraphics[width=\textwidth]{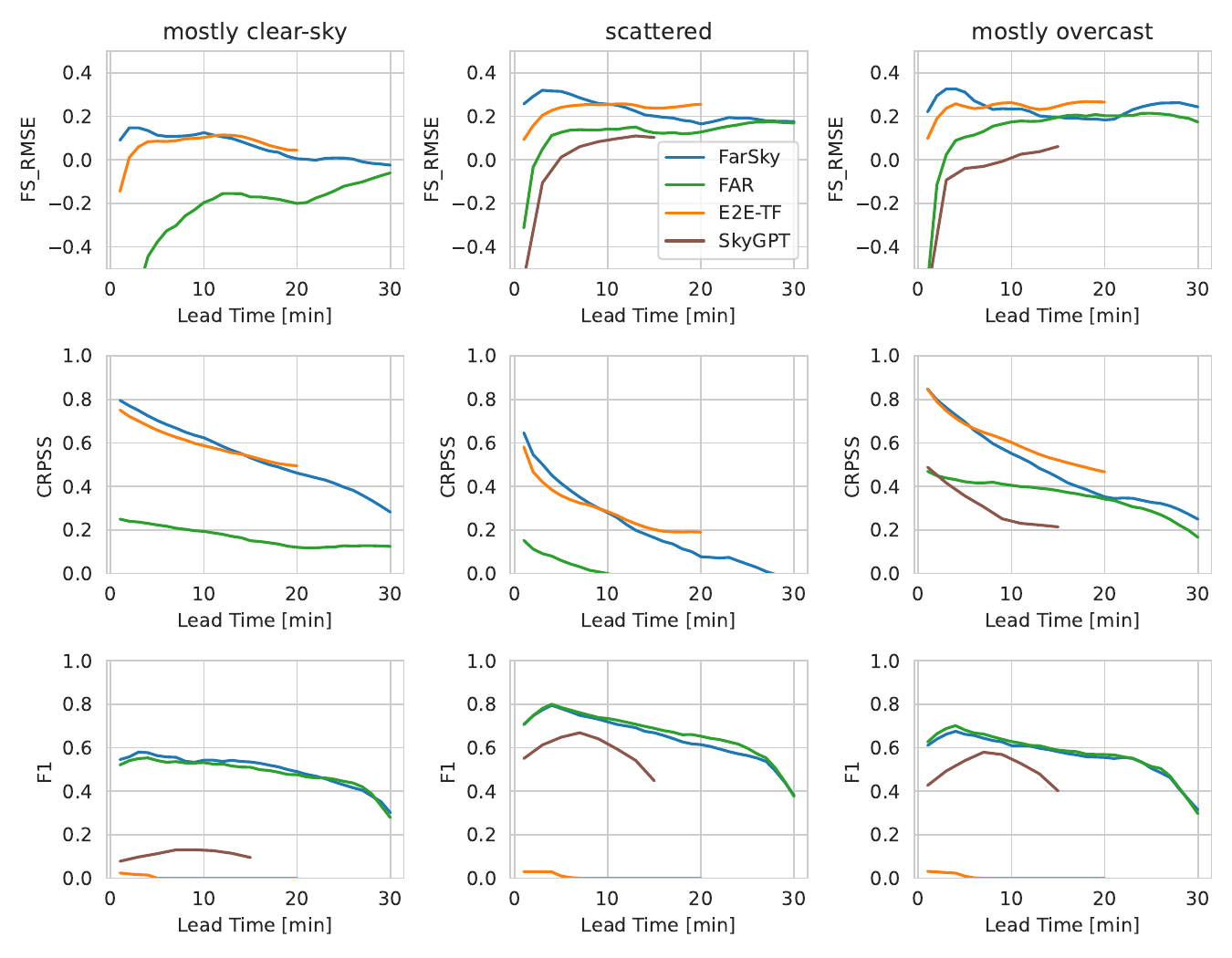}
\caption{FS, CRPSS, and ramp detection F1 score of the benchmark models as a function of forecast lead time for different atmospheric conditions on the \SVATest{} dataset. Columns correspond to mostly clear-sky, scattered-cloud, and mostly overcast conditions. Negative metric values are truncated to improve readability.}
\label{fig:metrics_sva_classes}
\end{figure}

Figures~\ref{fig:metrics_psa_classes} and~\ref{fig:metrics_sva_classes} compare deterministic, probabilistic, and ramp event performance under different atmospheric conditions derived from the DNI variability classes introduced in Section~\ref{sec:dni_var_classes}. 
The three considered regimes represent mostly clear-sky, scattered cloud, and mostly overcast conditions. 
Results are shown separately for both test datasets, \PSABench{} and \SVATest{}.

Under mostly clear-sky conditions, deterministic forecast skill remains comparatively low for all evaluated models. 
This behavior is expected, as persistence already provides highly accurate forecasts under nearly cloud-free conditions, leaving only limited potential for improvement. 
Nevertheless, FarSky consistently achieves the highest forecast skill at short lead times.
On the \PSABench{} dataset, the proposed approach outperforms the E2E-TF model throughout its entire forecasting horizon. 
This observation is noteworthy because the E2E-TF model additionally exploits recent irradiance measurements as input, allowing it to adapt its forecasts to the latest observations, whereas the generative approaches rely exclusively on all-sky imagery. 
On the \SVATest{} dataset, however, this advantage is limited to the first few lead times, after which increasing forecast bias reduces FarSky's forecast skill. 
A simple clear-sky post-processing strategy to mitigate this effect is investigated in Appendix~\ref{app:post_processing}.
Regarding SkyGPT, a substantially lower forecast skill is exhibited, remaining below the plotted range for all lead times, indicating that even for clear conditions the forecasts deviate strongly from the observed measurements.
The probabilistic evaluation reveals an almost complementary behavior.
CRPSS reaches its highest values under mostly clear-sky conditions, where forecast uncertainty is inherently small and the predictive distributions remain well concentrated. 
Here, FarSky and E2E-TF achieve very similar probabilistic forecast skill on both datasets, while the decoupled FAR pipeline performs noticeably worse. 
SkyGPT exhibits negative CRPSS on the \SVATest{} dataset, indicating that its probabilistic forecasts are less informative than the climatological reference. 
Visual inspection of representative forecasts suggests that the model frequently generates spurious cloud structures under otherwise clear-sky conditions, thereby introducing unnecessary forecast uncertainty. 
Consistent with the scarcity of rapid irradiance fluctuations under these conditions, ramp detection also represents the most challenging classification task.
Since only few ramp events occur, individual missed detections or false alarms have a comparatively large influence on the resulting classification metrics.

A different picture emerges under scattered cloud conditions. 
Here, all forecasting approaches achieve their highest forecast skill, reflecting the reduced performance of persistence during highly variable cloud situations. 
FarSky again provides the highest forecast skill at short lead times on both datasets and maintains this advantage over the complete forecasting horizon on the \PSABench{} dataset. 
On the \SVATest{} dataset, however, the E2E-TF model surpasses FarSky beyond approximately 10\,min.
This indicates that uncertainties in the generative prediction of future cloud evolution accumulate more rapidly at longer forecast horizons than the direct irradiance prediction errors of the E2E-TF model.
The probabilistic evaluation confirms that scattered cloud conditions constitute the most challenging regime for uncertainty estimation. 
CRPSS decreases considerably faster with lead time than for the other atmospheric conditions, reflecting the substantially increased uncertainty associated with rapidly evolving cloud fields. 
Nevertheless, FarSky retains the highest probabilistic skill among the generative approaches and shows a clear advantage over E2E-TF on the \PSABench{} dataset. 
Ramp event detection simultaneously reaches its highest F1 scores under scattered conditions, demonstrating that the diffusion-based forecasting approaches are able to capture the cloud dynamics responsible for pronounced irradiance variability particularly well.

The most interesting behavior is observed under mostly overcast conditions. 
Although FarSky again achieves the highest forecast skill at short lead times, its performance decreases noticeably beyond approximately 15\,min on the \PSABench{} dataset and eventually falls below that of the decoupled FAR pipeline. 
Together with the increasing positive mean bias discussed in the previous section, this behavior suggests that the generated cloud fields appear to become less extensive than those observed, resulting in an overestimation of irradiance at longer forecast horizons.
A plausible explanation is that predominantly overcast situations provide only limited information about the surrounding cloud field within the restricted field of view of the all-sky imager. 
Consequently, the diffusion model may increasingly converge towards more typical cloud configurations learned during training, which are dominated by scattered or partly clear conditions rather than persistent overcast cloud cover.
A similar trend is observed for CRPSS, where FarSky gradually loses its advantage over E2E-TF at longer lead times, particularly on the \SVATest{} dataset, suggesting that the deterioration in deterministic accuracy is accompanied by a reduction in probabilistic forecast skill.
In contrast, ramp detection performance remains comparatively stable under overcast conditions for the diffusion-based approaches, reaching F1 scores that fall between the ones at clear and scattered conditions.

Overall, the benchmark evaluation demonstrates that FarSky provides the strongest  forecasting performance, achieving the best balance across the considered deterministic, probabilistic, and ramp event evaluation metrics. 
In particular, the proposed framework closes the performance gap between generative forecasting and state-of-the-art MSE-optimized approaches while preserving the strong ramp event anticipation characteristic of generative video prediction.
The atmospheric-condition analysis further reveals complementary strengths and weaknesses of the evaluated forecasting paradigms. 
FarSky provides the highest deterministic forecast skill at short lead times across all considered cloud regimes and ranks among the strongest probabilistic forecasting approaches. 
Predominantly overcast conditions, however, emerge as the most challenging scenario for the proposed generative framework at longer forecast horizons.
At the same time, the results highlight that different evaluation targets emphasize different model properties. 
While latent-space coupling substantially improves absolute irradiance accuracy and probabilistic forecast skill, the strong ramp detection performance is primarily attributable to the underlying generative video prediction framework, which enables both FarSky and the decoupled baseline to capture relative irradiance changes effectively. 
Finally, noticeable performance differences remain between the \PSABench{} and the \SVATest{} datasets even after stratifying the evaluation by DNI variability classes. This suggests that the adopted classification provides only a coarse characterization of the prevailing atmospheric conditions.
Although it groups observations according to overall irradiance variability, substantial differences in cloud morphology and, in particular, the temporal evolution of cloud fields may still exist within the same variability class.
These findings further underline the importance of standardized benchmark datasets and comprehensive evaluation protocols for the fair comparison of solar forecasting methods.

\section{Conclusion}
\label{sec:conclusion}

This work introduced FarSky, a generative solar irradiance forecasting framework that combines latent-space video prediction with task-aware representation learning. 
By jointly optimizing image reconstruction and irradiance estimation within a shared latent space, the proposed DCAEIrr embeds irradiance-relevant information directly into the learned representations, enabling irradiance forecasting without relying on a separate regression model.

The experimental results demonstrate that latent-space coupling substantially improves the quality of the learned representations. 
Compared with a standalone irradiance regression model, the proposed multi-task representation learning approach significantly reduces irradiance estimation errors while maintaining nearly the same image reconstruction quality as the baseline autoencoder. 
These improvements carry over to the forecasting task, where FarSky consistently achieves lower deterministic errors, higher forecast skill, and better probabilistic performance than the corresponding decoupled pipeline. 
In particular, jointly learning irradiance-aware latent representations effectively mitigates the domain shift introduced by image reconstruction and video prediction, leading to a pronounced reduction in forecast bias.

Benchmark evaluations on the \PSABench{} and the \SVATest{} datasets further demonstrate the effectiveness of the proposed framework. 
Across both datasets, FarSky achieves the best overall balance of deterministic and probabilistic forecasting performance among the evaluated approaches, outperforming both the decoupled baseline and a state-of-the-art end-to-end Transformer model.
In particular, the proposed framework proves highly effective at predicting irradiance ramp events by separating cloud dynamics prediction from irradiance estimation. 
Explicitly predicting future cloud evolution enables the framework to capture the spatio-temporal variability underlying rapid irradiance changes, resulting in substantially better ramp detection than conventional deterministic forecasting approaches.
Consequently, latent-space coupling provides only limited additional benefits over the decoupled architecture for this particular task.

The atmospheric-condition analysis further reveals that the proposed framework is particularly effective under clear and scattered cloud conditions, while performance gains over the benchmark methods diminish under predominantly overcast conditions at longer lead times.
In addition, the probabilistic evaluation indicates that forecast uncertainty is slightly underestimated, suggesting that the limited number of generated stochastic trajectories constrains the calibration of predictive distributions.

Future work could therefore focus on improving uncertainty estimation by generating larger forecast ensembles or by incorporating dedicated probabilistic calibration techniques. 
Furthermore, extending the latent representation to include additional meteorological variables or leveraging larger and more diverse training datasets may improve robustness across different climatic regions. 
Finally, continued advances in diffusion-based video prediction, together with training on longer temporal contexts and forecast horizons, offer promising opportunities for further enhancing the performance of generative solar forecasting models.

Overall, the presented results demonstrate that integrating task-aware representation learning with generative video prediction provides an effective framework for short-term solar irradiance forecasting.
Beyond improving forecast accuracy, the proposed latent-space coupling highlights the potential of learning from multiple complementary tasks to encode physically meaningful information from all-sky imagery, representing a promising step toward more general-purpose vision models for atmospheric observations

\section*{Acknowledgments}
This research was funded by the German Federal Ministry for Economic Affairs and Energy through the PV-Reserve project (grant agreement number: 03EE1238A) based on a decision by the German Bundestag.

\appendix

\section{Implementation Details}
\label{sec:implementation_details}

This appendix summarizes the implementation details and hyperparameter configurations used throughout the experimental evaluation.
All models were trained on the common development dataset described in Section~\ref{sec:dataset_construction} and evaluated on the corresponding test sets.
Whenever possible, the original training procedures and hyperparameter settings of the respective baseline methods were adopted without modification.
Only deviations required for adapting the models to the considered forecasting task are reported in the main text.

\subsection{FarSky Configuration}
\label{sec:farsky_hyperparameters}

The proposed FarSky framework consists of the DCAEIrr representation model and the latent video prediction model FAR.
Tables~\ref{tab:dcaeirr_training_config} and~\ref{tab:far_training_config}
summarize the corresponding training configurations.

\begin{table}[!ht]
\centering
\caption{Training configuration of the enhanced autoencoder DCAEIrr.}
\label{tab:dcaeirr_training_config}
\begin{tabular}{ll}
\toprule
\textbf{Parameter} & \textbf{Value} \\
\midrule
Architecture & DCAEIrr \\
Input & Sky image \\
Target & Sky image, GHI \\
Image size & $128 \times 128$ pixels \\
Irradiance normalization & Division by solar constant \\
Irradiance prediction head & Global average pooling + two-layer MLP \\
Hidden dimension (MLP) & 128 \\
Activation function & SiLU \\
Output dimension & 1 (scalar irradiance) \\
Batch size & 8 per GPU \\
Context frames & 8 \\
Prediction frames & 8 \\
Input time range & $-15$\,min to $0$\,min \\
Prediction time range & $+1$\,min to $+15$\,min \\
Temporal resolution & 2\,min \\
Optimizer (generator/discriminator) & AdamW \\
Learning rate & $1 \times 10^{-4}$ \\
Weight decay & 0 \\
AdamW betas & $(0.5, 0.9)$ \\
Learning rate schedule & Constant \\
Warm-up iterations & 0 \\
Training iterations & 200{,}000 \\
EMA decay & 0.9999 \\
Maximum gradient norm & 1.0 \\
Reconstruction loss weight & 1.0 \\
Perceptual loss weight & 1.0 \\
Discriminator loss weight & 0.5 \\
Irradiance loss weight & 0.1 \\
Discriminator start iteration & 100{,}000 \\
\bottomrule
\end{tabular}
\end{table}

\begin{table}[!ht]
\centering
\caption{Training configuration of the latent diffusion-based FAR video prediction model.}
\label{tab:far_training_config}
\begin{tabular}{ll}
\toprule
\textbf{Parameter} & \textbf{Value} \\
\midrule
Architecture & FAR \\
Autoencoder & Frozen DCAEIrr \\
Training objective & Latent flow-matching objective \\
Loss function & MSE \\
Input image size & $128 \times 128$ pixels \\
Input & Latent sky representations \\
Target & Latent sky representations \\ 
Batch size & 16 per GPU \\
Context frames & 16 \\
Prediction frames & 16 \\
Input time range & $-15$,min to $0$,min \\
Prediction time range & $+1$,min to $+16$,min \\
Temporal resolution & 1\,min \\
Optimizer & AdamW \\
Learning rate & $1 \times 10^{-4}$ \\
Weight decay & 0 \\
AdamW betas & $(0.9, 0.999)$ \\
Learning rate schedule & Constant \\
Training iterations & 200{,}000 \\
EMA decay & 0.9999 \\
Maximum gradient norm & 1.0 \\
\bottomrule
\end{tabular}
\end{table}

\subsection{Decoupled Baseline Configuration}
\label{sec:decoupled_hyperparameters}
The decoupled baseline employs the original DCAE and FAR models. 
Since DCAEIrr differs from the original DCAE only by the addition of the irradiance prediction head, the image encoder and decoder were trained using the same hyperparameters and optimization procedure summarized in Table~\ref{tab:dcaeirr_training_config}. 
The FAR model was likewise trained using the configuration shown in Table~\ref{tab:far_training_config}. 
The image-based irradiance regressor constitutes an additional trainable component. 
Its training configuration is summarized in Table~\ref{tab:cnn_training_config}.

\begin{table}[!ht]
\centering
\caption{Training configuration of the image-based CNN irradiance regressor used in the decoupled baseline.}
\label{tab:cnn_training_config}
\begin{tabular}{ll}
\toprule
\textbf{Parameter} & \textbf{Value} \\
\midrule
Architecture & ResNet34 \\
Input & Sky image \\
Target & Clear-sky index of GHI \\
Image size & $128 \times 128$ pixels \\
Batch size & 64 \\
Optimizer & AdamW \\
Learning rate & $5 \times 10^{-5}$ \\
Weight decay & 0 \\
Learning rate schedule & One-cycle \\
Loss function & Mean squared error (MSE) \\
Training iterations & 200{,}000 \\
Early stopping & Enabled \\
\bottomrule
\end{tabular}
\end{table}

\section{Post-Processing for Clear-sky Conditions}
\label{app:post_processing}
\begin{figure}[htbp]
\centering
\includegraphics[width=\textwidth]{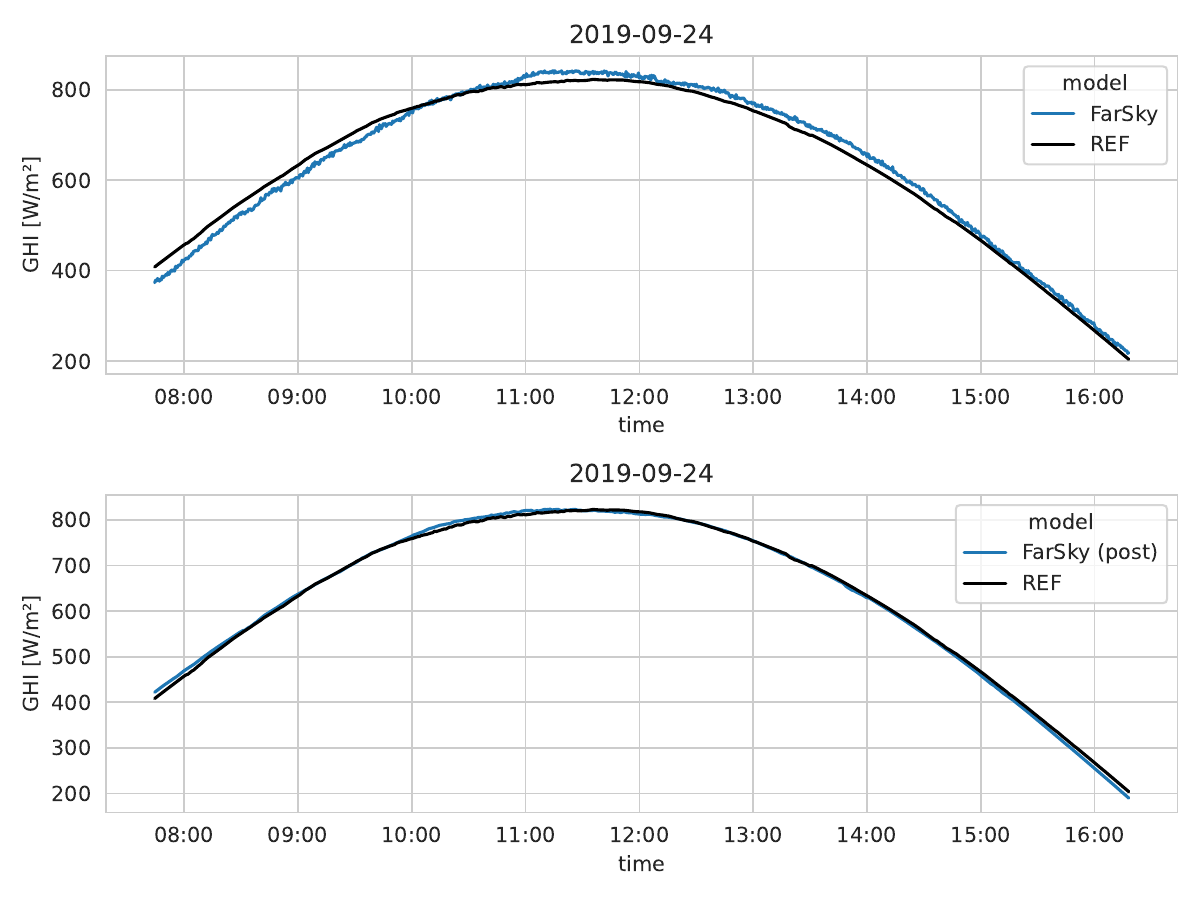}
\caption{Exemplary 30\,min GHI forecasts of FarSky for a clear-sky day from the \PSABench{} dataset. Top: original FarSky forecast. Bottom: forecast after applying the proposed clear-sky post-processing.}
\label{fig:farsky_post_forecast_clear}
\end{figure}

The proposed framework FarSky deliberately relies solely on all-sky imagery and therefore does not incorporate an explicit self-calibration mechanism based on recent irradiance measurements.
Consequently, small systematic deviations between the predicted and measured irradiance may accumulate over time. 
Combined with the slight positive bias observed for the generative forecasts at longer lead times, this can result in a minor but noticeable offset between forecast and measurement curves under persistent clear-sky conditions, as illustrated in the upper panel of Figure~\ref{fig:farsky_post_forecast_clear}.

For operational scenarios where real-time GHI measurements are available, this effect can be mitigated through a simple post-processing step. 
The proposed approach employs the clear-sky detection algorithm of~\cite{Reno2016_Identification}, using an adapted implementation provided in~\cite{Anderson2023_pvlib}. 
The algorithm identifies periods of clear-sky conditions within a configurable temporal window. 
For each window, the measured GHI is compared against the corresponding clear-sky irradiance using a set of statistical consistency tests, including the mean and maximum irradiance, the signal line length, and the variability of the irradiance slope. 
A window is classified as clear only if all criteria satisfy predefined thresholds. 
In contrast to the original formulation, which evaluates measurements before and after the timestamp of interest, only historical observations are considered here to avoid information leakage from future measurements. 
Whenever clear-sky conditions are detected, the corresponding FarSky forecasts are replaced by the Scaled Persistence forecast, which is expected to be optimal under such conditions. 
The resulting forecasts for an exemplary clear-sky day are shown in the lower panel of Figure~\ref{fig:farsky_post_forecast_clear}.

\begin{figure}[htbp]
\centering
\includegraphics[width=\textwidth]{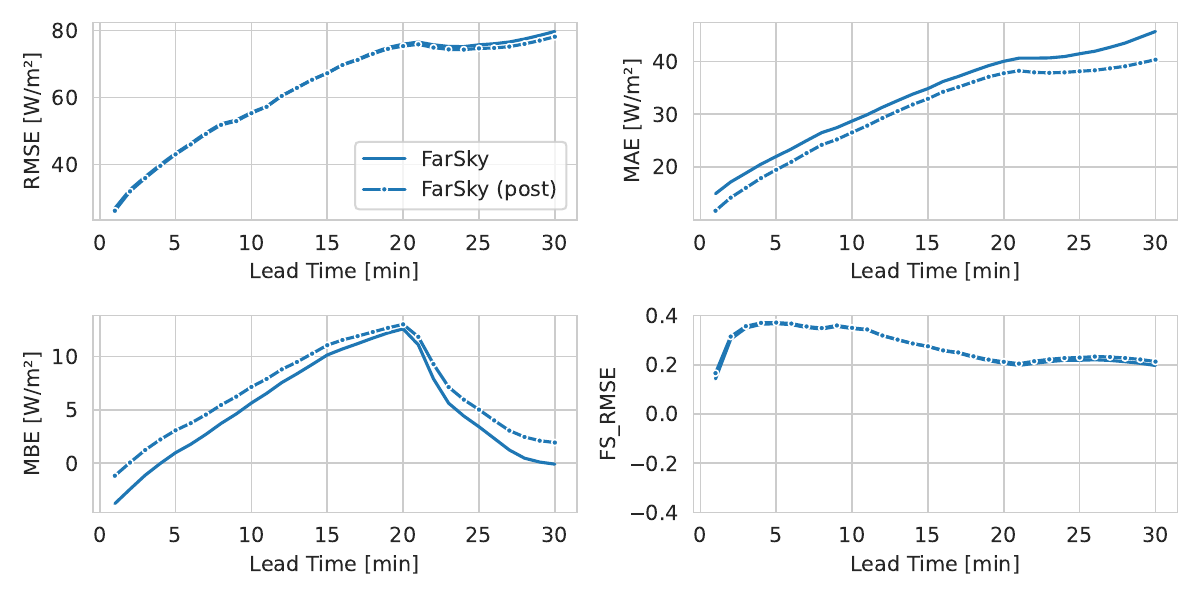}
\caption{Deterministic forecasting performance of FarSky before and after clear-sky post-processing on the \PSABench{} dataset as a function of lead time.}
\label{fig:farsky_post_det_metrics_per_lead_time}
\end{figure}

The influence of the post-processing on the deterministic forecasting metrics over all evaluation samples is summarized in Figure~\ref{fig:farsky_post_det_metrics_per_lead_time}. 
Since clear-sky periods already exhibit comparatively small forecast errors, replacing the original forecasts affects the overall RMSE and forecast skill only marginally. 
The largest improvement is observed for the MAE, which is reduced by approximately 2 to 5\,W\,m$^{-2}$ across the forecasting horizon for the \PSABench{} dataset.
Note that such improvements depend a lot on the fraction of clear-sky periods within the evaluated dataset.

An analysis restricted to the detected clear-sky periods is provided in Table~\ref{tab:farsky_post_det_metrics_clear}. 
Under these conditions, forecast errors are generally much smaller than for the complete dataset, and the proposed post-processing further reduces both RMSE and MAE while nearly eliminating the systematic bias. 
The forecast skill relative to Scaled Persistence becomes zero, as expected, because the FarSky forecasts are replaced by the persistence prediction whenever clear-sky conditions are detected. 
Interestingly, while the post-processing improves the bias during clear-sky periods, it increases the overall MBE on the complete evaluation dataset, as shown in Figure~\ref{fig:farsky_post_det_metrics_per_lead_time}. 
This effect arises because FarSky exhibits a slight negative bias during clear-sky conditions but tends to overestimate irradiance at longer lead times overall. 
Replacing the negatively biased clear-sky forecasts with the nearly unbiased Scaled Persistence forecasts therefore removes part of the negative contribution to the global MBE, causing the average bias over the complete dataset to shift further towards positive values.

\begin{table}[t]
\centering
\caption{Average performance over the entire forecast horizon of 30\,min of FarSky on detected clear sky periods compared to Scaled Persistence (SP).}
\label{tab:farsky_post_det_metrics_clear}
\begin{tabular}{lrrrr}
\toprule
Model & RMSE & MAE & MBE & FS$_{\mathrm{RMSE}}$ \\
\midrule
FarSky & 13.02 & 8.00 & -3.42 & -0.439 \\
SP & 10.30 & 2.64 & 0.49 & 0.000 \\
\bottomrule
\end{tabular}
\end{table}

\bibliographystyle{IEEEtran} 
\bibliography{literature}

\end{document}